%% file: main.tex
\documentclass{article}

\usepackage{arxiv}

\usepackage[utf8]{inputenc}
\usepackage[T1]{fontenc}
\usepackage{hyperref}
\usepackage{url}
\usepackage{booktabs}
\usepackage{amsfonts}
\usepackage{amsmath}
\usepackage{amssymb}
\usepackage{microtype}
\usepackage{graphicx}
\usepackage{caption}
\usepackage{multirow}
\usepackage{algorithm}
\usepackage{algpseudocode}
\usepackage{xcolor}
\usepackage[numbers,sort&compress]{natbib}

\newcommand{\revnew}[1]{{\color{red}#1}}

\title{Optimal Sequential Decision-Making for Error Propagation Mitigation in Digital Twins}

\author{%
  Annice Najafi \\
  Department of Industrial and Manufacturing Engineering\\
  California State Polytechnic University, Pomona\\
  Pomona, CA, 91768, USA\\
  \texttt{annicenajafi27@gmail.com}
  \And
  Shokoufeh Mirzaei\thanks{Corresponding author.} \\
  Department of Industrial and Manufacturing Engineering\\
  California State Polytechnic University, Pomona\\
  Pomona, CA, 91768, USA\\
  \texttt{smirzaei@cpp.edu}
}

\begin{document}
\maketitle

\begin{abstract}
Here, we explore the problem of error propagation mitigation in modular digital twins as a sequential decision process. Building on a companion study that used a Hidden Markov Model (HMM) to infer latent error regimes from surrogate-physics residuals, we develop a Markov Decision Process (MDP) in which the inferred regimes serve as states, corrective interventions serve as actions, and a scalar reward that takes into consideration the cost-benefit tradeoff between system fidelity and maintenance expense. The baseline transition matrix is extracted from the HMM-learned parameters. We then extend the formulation to a Partially Observable MDP (POMDP) that accounts for the imperfect nature of regime classification by maintaining a belief distribution updated via Bayesian filtering, with the HMM confusion matrix serving as the observation model. Both formulations are solved via dynamic programming and validated through Gillespie stochastic simulation. We then benchmark two model-free reinforcement learning algorithms, Q-learning and REINFORCE, to assess whether effective policies can be learned without explicit model knowledge. A systematic comparison of different intervention policies (optimal MDP, POMDP, Q-learning, REINFORCE, a $k$-step consecutive-observation reactive baseline, MCDA/TOPSIS, and no intervention) demonstrates that the MDP policy achieves the highest cumulative reward and fraction of time in nominal operation, while the POMDP recovers approximately 95\% of MDP performance under realistic observation noise. Sensitivity analyses across observation quality, repair probability, and discount factor confirm the robustness of these conclusions, and the major gaps in the policy hierarchy are statistically significant at $p < 0.001$. The gap between MDP and POMDP performance quantifies the value of information thereby providing a principled criterion for investing in improved classification accuracy.
\end{abstract}


\input{body}

\bibliographystyle{unsrtnat}
\bibliography{references}

\end{document}

%% file: body.tex
\section{Introduction}
\label{sec:introduction}

Digital twins, which are virtual models of physical systems that are updated with data from their real-world counterparts, are increasingly used in engineering for monitoring, diagnosis, prediction, and control \cite{singh2021digital, sharma2022digital}. Their value is extremely important in helping to make decisions before problems worsen or performance declines \cite{sharma2022digital, agrell2023optimal}. In a modular design, the output of one sub-model becomes the input for another. Therefore, even a small error introduced early on can spread through later modules and eventually affect system-level estimates, health indicators, and control suggestions \cite{najafi2026hmm}. As a result, reducing error propagation is extremely important for reliable digital twin use, especially when the twin is used to support operational or maintenance decisions \cite{agrell2023optimal, palotai2025surrogate}.

This issue is especially significant within modular digital twins constructed using surrogate models \cite{palotai2025surrogate, desai2023enhanced}. Surrogates such as AutoRegressive eXogenous (ARX) models \cite{ljung1999system} offer computational efficiency and scalability, but they are inherently approximate and may degrade when the underlying physical process changes, when sensor characteristics drift, or when noise levels increase \cite{palotai2025surrogate, desai2023enhanced}. In such systems, discrepancies between the physical process and its surrogate representation do not remain localized \cite{palotai2025surrogate, desai2023enhanced}. Rather, they alter downstream residuals and interfere with state estimation \cite{lin2026explainable, najafi2026hmm}. A dependable digital twin must therefore identify the latent operating regime underlying inaccurate outputs and decide how to intervene so that fidelity is restored at minimal cost \cite{agrell2023optimal, ghosh2019hidden}. In our previous work, we addressed the first part of this problem by introducing a HMM framework for inferring latent error regimes from residual features extracted at module boundaries. We showed that distinct regimes such as nominal behavior, sensor-noisy operation, plant-dynamics mismatch, and drift can be recovered from residual data with useful accuracy, thereby providing a probabilistic diagnostic layer for modular digital twins \cite{najafi2026hmm}. Corrective actions were then selected using Technique for Order Preference by Similarity to Ideal Solution (TOPSIS) as a multi-criteria ranking heuristic \cite{najafi2026hmm, hwang1981multiple, najafi2025rmcda}. While that approach demonstrated regime-aware mitigation is feasible, it remained fundamentally myopic. TOPSIS evaluates actions independently at each decision epoch and does not model the stochastic evolution of error regimes over time \cite{hwang1981multiple}, the downstream consequences of present actions, or the possibility that additional information may improve future decisions. These limitations are especially important when interventions are costly, when regimes are persistent, and when misclassification can lead to unnecessary or mistargeted repairs.

Sequential decision-making under uncertainty is naturally addressed through Markov Decision Processes (MDP) and Partially Observable Markov Decision Processes (POMDP) \cite{puterman2014markov, kaelbling1998planning, smallwood1973pomdp}. In an MDP, the decision-maker selects actions based on the current state while accounting for long-run discounted consequences through the transition structure and reward function \cite{puterman2014markov}. In a POMDP, the decision-maker instead acts on a belief state when the true underlying state is not directly observable \cite{smallwood1973pomdp, kaelbling1998planning}. These frameworks are well suited to maintenance optimization, structural health monitoring, and other engineering problems in which intervention must balance performance recovery against cost \cite{papakonstantinou2014planninga, papakonstantinou2014planningb, arcieri2023bridging}. They are especially appropriate for digital twins, where latent degradation regimes evolve over time and the available information is typically indirect and noisy. For example, Agrell et al.\ framed probabilistic digital twins as a basis for sequential decision-making under uncertainty and discussed MDP/POMDP formulations, however, they did not address the specific problem of mitigation of error propagation in modular digital twins \cite{agrell2023optimal}.

Thus the problem of mitigating error propagation in modular digital twins has not, to our knowledge, been formalized as a sequential decision process whose transition and observation models are derived from data-driven latent-regime inference. Existing digital-twin mitigation strategies are often heuristic, threshold-based, or single-step in nature, and thus do not explicitly quantify the value of planning ahead or the cost of imperfect observability \cite{agrell2023optimal}. A digital twin deployed in practice must operate under uncertainty not only about the physical system but also about the quality of its own internal representations \cite{agrell2023optimal, palotai2025surrogate}. A framework that can consider regime persistence, intervention efficacy, classification uncertainty, and long-run reward is therefore needed if digital twins are to become reliable decision-support systems rather than descriptive monitoring tools \cite{agrell2023optimal, puterman2014markov, kaelbling1998planning}.

Here, we formulate error propagation mitigation in modular digital twins first as an MDP and then as a POMDP. In the proposed formulation, the latent error regimes inferred from the HMM define the state space, corrective interventions define the action space, and a reward function encodes the tradeoff between preserving system fidelity and avoiding unnecessary maintenance cost. A distinguishing feature of the framework is that its baseline transition dynamics are not imposed heuristically; instead, they are extracted directly from the HMM-learned transition parameters, thereby grounding the sequential decision model in data. We then extend the formulation to the partially observable case by using the empirical HMM confusion matrix as the observation model and allow the controller to maintain and update a belief distribution over regimes rather than relying on hard classifications alone. Residuals drive regime inference which in turn determines the sequential intervention policy.

The results show that this shift is consequential. The MDP provides the upper bound achievable under perfect regime knowledge, while the POMDP captures the more realistic case in which the true regime is only indirectly observed. By comparing these two frameworks with model-free learning methods, heuristic MCDA, and no intervention, we show how much performance is lost to myopia, how much is lost to partial observability, and how much can be recovered by maintaining a belief state and planning sequentially.

\section{Methods}
\label{sec:methods}

Let the digital twin under consideration be a modular pipeline of $n$ interconnected modules $M_1, M_2, \ldots, M_n$. For each module $M_i$, an input $x_i$ is mapped to an output $y_i$, and the output of each module is fed as input into the next, so signals propagate sequentially down the chain. The physical asset produces ground-truth signals $y^{\text{phys}}_i$ at every module boundary. In parallel, the digital twin approximates each module's behavior with a surrogate model $\hat{f}_i$, whose outputs constitute the twin's predictions $y^{\text{surr}}_i$. Surrogates are not restricted to any particular class and may be chosen to suit the application:
\begin{enumerate}
      \item data-driven representations such as ARX models, neural
      networks, or Gaussian processes;
      \item reduced-order formulations derived from the underlying
      physics;
      \item hybrid constructions combining physical priors with learned
      components.
\end{enumerate}

During operation, discrepancies between the physical system outputs and their digital counterparts emerge as module-level residuals whose statistical signatures change as the system transitions among latent error regimes. The inference layer described in our companion paper \cite{najafi2026hmm} uses a Gaussian-emission HMM to decode these residual streams into a discrete regime label at every time step.

The decision problem that arises naturally is that given the current error regime, which corrective intervention, if any, should be applied so as to steer the system back toward nominal operation while avoiding unnecessary maintenance costs? We formalize this sequential decision problem as a POMDP, where error regimes serve as states, corrective interventions serve as actions, and a scalar reward encodes the cost-benefit tradeoff between system fidelity and intervention expense. First, let us explain a simpler and more ideal scenario where the state of the system is observable similar to our presented Markov model framework in Najafi et al.\ and add a decision layer \cite{najafi2025stochastic}.

\subsection{MDP Definition}
\label{sec:mdp_def}

An MDP is defined by the tuple $(\mathcal{S}, \mathcal{A}, T, R, \gamma)$, where $\mathcal{S}$ is the finite state space, $\mathcal{A}$ is the finite action space, $T : \mathcal{S} \times \mathcal{A} \times \mathcal{S} \to [0,1]$ is the transition function with $T(s, a, s') = \Pr(s_{t+1} = s' \mid s_t = s, a_t = a)$, $R : \mathcal{A} \times \mathcal{S} \to \mathbb{R}$ is the reward function, and $\gamma \in [0,1)$ is the discount factor. The objective is to find a stationary policy $\pi^* : \mathcal{S} \to \mathcal{A}$ that maximizes the expected infinite-horizon discounted return:
\begin{equation}
\pi^* = \arg\max_{\pi}\; \mathbb{E}\left[\sum_{t=0}^{\infty} \gamma^t R(\pi(s_t), s_t)\right].
\label{eq:mdp_objective}
\end{equation}

\subsection{Transition Model}
\label{sec:transition_model}

The transition model is decomposed into a baseline component in order to capture the natural evolution of error regimes in the absence of intervention, and an action-dependent component that models the corrective effect of each intervention. The baseline transition matrix $A^{(\text{NoAction})} \in \mathbb{R}^{K \times K}$ captures the natural state evolution in the absence of corrective action; its calibration from data is described in Section~\ref{sec:case_study}. Each corrective action $a$ is associated with a repair probability $\rho_{a,s} \in [0,1]$ representing the probability that applying action $a$ while the system is in state $s$ successfully transitions the system to the healthy state in the next step (specific values in Table~\ref{tab:repair} for the case study presented in this article).

\subsubsection{Action-Dependent Transition Matrices}

Given the baseline matrix and repair probabilities, the action-dependent transition matrix $A^{(a)}$ is constructed row-wise. For each state $s$:
\begin{equation}
A^{(a)}_{s,s'} = (1 - \rho_{a,s})\, A^{(\text{NoAction})}_{s,s'} + \rho_{a,s}\, \mathbf{I}[s' = \textsc{Nominal}],
\label{eq:action_transition}
\end{equation}
where $\mathbf{I}$ is the indicator function. When $\rho_{a,s} = 0$, the row reduces to the baseline. When $\rho_{a,s} > 0$, a fraction $\rho_{a,s}$ of the transition probability mass is redirected to \textsc{Nominal}, while the remaining fraction $(1 - \rho_{a,s})$ retains the baseline dynamics. We note that by construction, each row of $A^{(a)}$ sums to $1$ and preserves the stochastic matrix property.

\subsection{Reward Function}
\label{sec:reward}

The reward function $R(a, s)$ assigns a scalar payoff for executing action $a$ in state $s$ and encodes both the benefit of a correct intervention and the cost of an unnecessary or mismatched one.

\subsection{Value Iteration}
\label{sec:value_iteration}

The optimal value function $V^*(s)$ and policy $\pi^*(s)$ are obtained via value iteration \cite{bellman1957dp}. The Bellman optimality equation is:
\begin{equation}
V^*(s) = \max_{a \in \mathcal{A}} \left[ R(a, s) + \gamma \sum_{s' \in \mathcal{S}} A^{(a)}_{s,s'}\, V^*(s') \right],
\label{eq:bellman}
\end{equation}
with the optimal action in each state given by:
\begin{equation}
\pi^*(s) = \arg\max_{a \in \mathcal{A}} \left[ R(a, s) + \gamma \sum_{s' \in \mathcal{S}} A^{(a)}_{s,s'}\, V^*(s') \right].
\label{eq:optimal_policy}
\end{equation}
Equivalently, the action-value function $Q^{(n)}(a, s)$ at iteration $n$ is:
\begin{equation}
Q^{(n)}(a, s) = R(a, s) + \gamma \sum_{s'} A^{(a)}_{s,s'}\, V^{(n)}(s'),
\label{eq:q_function}
\end{equation}
from which $V^{(n+1)}(s) = \max_a Q^{(n)}(a, s)$. Value iteration is initialized with base case $V^{(0)}(s) = 0$ for all $s \in \mathcal{S}$ and iterated until $\|V^{(n+1)} - V^{(n)}\|_\infty$ falls below a prescribed tolerance. The specific parameter settings and convergence results for the case study are reported in Section~\ref{sec:case_study}.

\subsection{Continuous-Time Embedding}
\label{sec:ctmc}

While value iteration operates in discrete time, the system's physical dynamics typically evolve in continuous time. To allow for stochastic simulation and to obtain a closed-form theoretical value function for validation, we embed the discrete-time MDP into a continuous-time Markov chain (CTMC) framework \cite{puterman2014markov}.

\subsubsection{Rate Matrices}

For each action $a \in \mathcal{A}$, we approximate the continuous-time rate (generator) matrix via the first-order embedding
\begin{equation}
Q^{(a)} \approx \frac{A^{(a)} - I}{\Delta t},
\label{eq:rate_matrix}
\end{equation}
where the off-diagonal entries $Q^{(a)}_{s,s'} \geq 0$ ($s \neq s'$) are transition rates (s$^{-1}$) and each row sums to zero. The exact continuous-time embedding requires $Q^{(a)} = \log(A^{(a)}) / \Delta t$, where $\log$ denotes the matrix logarithm. The first-order approximation is exact only when $A^{(a)} = I$ and accumulates $O(\|A^{(a)} - I\|^2)$ truncation error otherwise: small for the baseline $A^{(\text{NoAction})}$ rows, whose deviation from the identity is at most ${\approx} 0.14$, but larger for action-modified rows where a repair redirects 40\% to 80\% of the probability mass to \textsc{Nominal}. Both the Gillespie simulation and the closed-form Bellman solution use the same approximate $Q^{(a)}$, so their agreement certifies internal consistency of the continuous-time validation pipeline rather than exact equivalence to the original discrete-time MDP.

\subsubsection{Continuous Discount Rate and Reward Rates}

The discrete discount factor $\gamma$ per step corresponds to a continuous discount rate $\rho$ and continuous reward rate $c(s, a)$:
\begin{equation}
\rho = -\frac{\ln \gamma}{\Delta t}, \qquad c(s, a) = \frac{R(a, s)}{\Delta t}.
\label{eq:continuous_rates}
\end{equation}
The specific numerical values for the case study are reported in Section~\ref{sec:case_study}.

\subsubsection{Continuous-Time Bellman Equation}

Under a fixed stationary policy $\pi$, define the policy-specific rate matrix $Q_\pi$ by $[Q_\pi]_{s,s'} = [Q^{(\pi(s))}]_{s,s'}$ and the policy reward-rate vector $c_\pi$ by $[c_\pi]_s = c(s, \pi(s))$. The continuous-time value function $V_{\text{ct}}(s)$ satisfies:
\begin{equation}
(\rho I - Q_\pi)\, V_{\text{ct}} = c_\pi,
\label{eq:ct_bellman}
\end{equation}
with unique solution $V_{\text{ct}} = (\rho I - Q_\pi)^{-1}\, c_\pi$. We note that invertibility is guaranteed because $\rho > 0$ ensures all eigenvalues of $Q_\pi - \rho I$ have strictly negative real part.

\subsection{Gillespie Simulation}
\label{sec:gillespie}

We simulate the continuous-time MDP with the Gillespie Stochastic Simulation Algorithm \cite{gillespie1977ssa}, which produces exact sample paths of the CTMC. Starting from state $s_0$ at $t = 0$, each iteration of the algorithm proceeds as follows. Given the current state $s$, we select the action $a = \pi(s)$, extract the nonnegative off-diagonal rates $\lambda_{s'} = \max(Q^{(a)}_{s,s'},\, 0)$, and form the total exit rate $\Lambda = \sum_{s' \neq s} \lambda_{s'}$. We then draw a sojourn time $\tau \sim \text{Exp}(\Lambda)$. If $t + \tau > T_{\text{end}}$ the trajectory terminates; otherwise the next state is sampled from $\text{Categorical}(\lambda_1/\Lambda, \ldots, \lambda_K/\Lambda)$ and the clock advances to $t + \tau$. The procedure repeats until the horizon is reached.

The discounted return along a trajectory with $N$ events at times $t_0 < t_1 < \cdots < t_N$ and states $s_0, s_1, \ldots, s_N$ is computed exactly via the closed-form segment integral:
\begin{equation}
J = \sum_{i=0}^{N} \frac{c(s_i, \pi(s_i))}{\rho}\, e^{-\rho t_i}\, (1 - e^{-\rho \Delta_i}),
\label{eq:segment_integral}
\end{equation}
where $\Delta_i = t_{i+1} - t_i$ and $t_{N+1} \equiv T_{\text{end}}$. The theoretical expected cumulative reward curve for comparison is:
\begin{equation}
J_{\text{th}}(t) = b_0^\top V_{\text{ct}} - b_0^\top \exp[(Q_\pi - \rho I)\, t]\, V_{\text{ct}},
\label{eq:theory_curve}
\end{equation}
evaluated numerically on a fine grid ($\Delta t = 0.01$\,s) using the matrix exponential.

\subsection{Extension to a Semi-Markov Decision Process}
\label{sec:smdp}

The MDP formulation in Section~\ref{sec:mdp_def} assumes decisions are made on a fixed discrete grid with step $\Delta t$, and the continuous-time embedding of Section~\ref{sec:ctmc} assumes that sojourn times are exponentially distributed and that actions take effect instantaneously. Both assumptions can be relaxed by promoting the MDP to a semi-Markov decision process (SMDP)~\cite{puterman2014markov, bradtke1995smdp}. In an SMDP the kernel $T(s, a, s') = \Pr(s_{t+1} = s' \mid s_t = s, a_t = a)$ is replaced by a joint distribution over next state and holding time,
\begin{equation}
P(s',\, \tau \mid s,\, a) \;=\; \Pr(s_{t+1} = s',\, \tau_{t+1} - \tau_t \in [\tau,\, \tau + d\tau] \mid s_t = s,\, a_t = a),
\label{eq:smdp_kernel}
\end{equation}
and the discounted Bellman equation generalizes to
\begin{equation}
V^*(s) = \max_{a \in \mathcal{A}}\, \mathbb{E}_{s',\tau \sim P(\cdot \mid s, a)}\!\left[\, \int_{0}^{\tau}\! e^{-\rho u}\, c(s, a)\, \mathrm{d}u \;+\; e^{-\rho \tau}\, V^*(s')\,\right].
\label{eq:smdp_bellman}
\end{equation}
Our current model is essentially the special case in which holding times are geometric with rate $1/\Delta t$ (alternatively, one could say $\tau$ is exponentially distributed with rate $|Q^{(a)}_{s,s}|$ in the continuous-time embedding) and the next-state and holding-time distributions factorize as $P(s', \tau \mid s, a) = T(s, a, s') f_\tau(\tau \mid s, a)$. In this sense the Gillespie simulator of Section~\ref{sec:gillespie} is already operating on an SMDP sample path whose kernel happens to be exponential. Replacing $\text{Exp}(\Lambda)$ by an empirically measured holding-time distribution $f_\tau$ would let the framework absorb non-exponential action-execution delays (e.g., the fixed recalibration window for \textsc{ReidentPlant} or the ramp-up time for \textsc{IncFilter}) with no change to the decision layer. We return to this extension in Section~\ref{sec:limitations}.

\subsection{Extended POMDP Model}
\label{sec:pomdp}

The MDP formulation assumes perfect knowledge of the current error regime at every time step. In practice, the true regime is never directly observable. As modeled in our previous paper the agent perceives the system state only through classifications produced by the HMM, which are subject to misidentification errors \cite{najafi2026hmm}. For example, in the specific case of our previous article the \textsc{SensorNoisy} regime is particularly susceptible to confusion with \textsc{Nominal} because elevated sensor variance does not produce a large mean shift in the residual features. An agent that treats HMM labels as ground truth may select \textsc{NoAction} when the system has actually entered a noisy-sensor condition, degrading performance relative to the MDP bound. A principled treatment of observation uncertainty is therefore required.

\subsubsection{POMDP Formulation}
\label{sec:pomdp_def}

\begin{figure*}[htbp]
    \centering
    \includegraphics[width=1.1\linewidth]{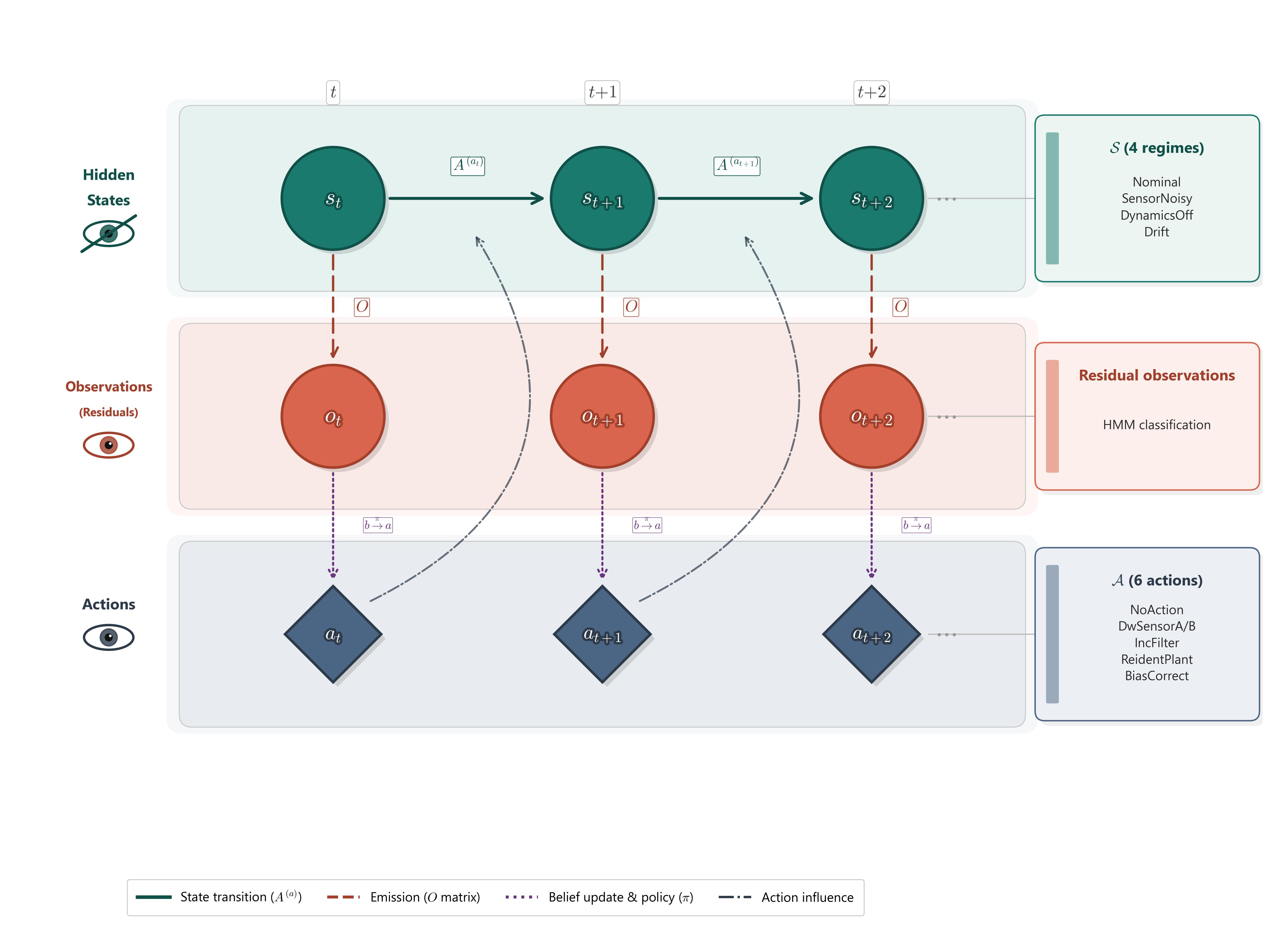}
    \caption{Graphical model of the POMDP framework for error propagation mitigation. The top layer represents the four latent error regimes (hidden states), which evolve according to action-dependent transition matrices $A^{(a)}$. The middle layer represents the observations generated by HMM classification of surrogate-physics residuals, linked to the hidden states through the observation matrix $O$. The bottom layer represents the corrective actions selected by the POMDP policy $\pi(b)$, which maps the current belief distribution $b$ over regimes to an intervention. Crossed and open eye symbols indicate unobservable and observable quantities, respectively. At each time step, the agent receives a noisy observation, updates its belief via Bayesian filtering, selects an action, and influences the next state transition.}
    \label{POMDP_Graph}
\end{figure*}

We extend the MDP tuple to a POMDP defined by $(\mathcal{S}, \mathcal{A}, T, R, \gamma, \mathcal{Z}, O, b_0)$, where $\mathcal{Z} = \{z_1, \ldots, z_K\}$ is the observation space (the same regime labels received through a noisy channel), $O : \mathcal{S} \times \mathcal{Z} \to [0,1]$ is the observation model, and $b_0 \in \Delta^{K-1}$ is the initial belief. In our setting, $\mathcal{Z} = \mathcal{S}$: the observations are regime labels, but these labels are noisy reflections of the true underlying regime rather than the regime itself. Rather than committing to a single regime label, the agent now maintains a belief state $b \in \Delta^{K-1}$, a probability distribution over all four regimes, updated via Bayesian filtering after each action-observation pair. We provide a graphical representation of this framework in Figure~\ref{POMDP_Graph}.

\subsubsection{Observation Model}
\label{sec:observation_model}

The observation matrix $O \in \mathbb{R}^{K \times K}$ with $O_{s,z} = P(\text{observe } z \mid \text{true state } s)$ is derived from the confusion matrix of the classifier evaluated against ground-truth labels. Each row sums to unity. The specific observation matrix used in this study, including the asymmetric classification accuracy across regimes, is described in Section~\ref{sec:case_study}.

\subsubsection{Belief Update}
\label{sec:belief_update}

The agent maintains a belief state $b \in \Delta^{K-1}$, a probability distribution over $\mathcal{S}$. After taking action $a$ and receiving observation $z$, the belief updates via:
\begin{equation}
b'(s') = \eta\, O(s', z)\, \sum_{s} A^{(a)}_{s,s'}\, b(s),
\label{eq:belief_update}
\end{equation}
where $\eta$ is a normalizing constant that ensures $\sum_{s'} b'(s') = 1$. In the continuous-time Gillespie setting, the prediction step uses the matrix exponential $\exp(Q^{(a)} \tau)$ to propagate the belief over the sojourn interval $\tau$, followed by the Bayesian observation update at the transition event. If the normalization denominator underflows to zero (an extremely rare event caused by an observation that is effectively impossible under the current belief), the belief is reset to the uniform distribution $b = (1/K, \ldots, 1/K)$.

\subsubsection{Point-Based Value Iteration}
\label{sec:pbvi}

We adopt Point-Based Value Iteration (PBVI) \cite{pineau2003pbvi}, which restricts Bellman iterations to a finite set of sampled belief points rather than computing the exact solution over the entire continuous belief simplex. We represent the value function as the upper envelope of $\alpha$-vectors:
\begin{equation}
V(b) = \max_{\alpha \in \Gamma}\; \alpha \cdot b,
\label{eq:alpha_value}
\end{equation}
where $\Gamma$ is the set of $\alpha$-vectors maintained by the solver. Each $\alpha$-vector is a $K$-dimensional vector that encodes the value of a particular action choice contingent on the true state.

Belief points are generated by combining $K$ corner beliefs (pure states), $\binom{K}{2}$ edge midpoints, $\binom{K}{3}$ face centroids, the simplex centroid $(1/K, \ldots, 1/K)$, and random Dirichlet$(1, \ldots, 1)$ samples. Using this diverse set we ensure the coverage of the belief simplex including its boundaries. The $\alpha$-vector set is initialized using the MDP value function $V^*_{\text{MDP}}(s)$ as a warm start, which provides PBVI with a near-optimal starting point and substantially accelerates convergence. Convergence is declared by policy stability on the belief grid where the greedy action associated with each sampled belief is checked after every iteration, and iteration comes to a stop when this mapping is unchanged for a fixed number of consecutive sweeps or when the budget is exhausted, whichever occurs first (the residual norm is reported for diagnostic purposes but is not used as a stopping criterion, for the reasons discussed in Section~\ref{sec:case_study}).

At runtime, the POMDP policy selects the action associated with the $\alpha$-vector that achieves the highest inner product with the current belief:
\begin{equation}
\pi_{\text{POMDP}}(b) = \text{action}\!\left(\arg\max_{\alpha \in \Gamma}\; \alpha \cdot b\right).
\label{eq:pomdp_policy}
\end{equation}

\subsubsection{Value of Information}
\label{sec:voi}

The gap between MDP and POMDP expected returns quantifies the cost of partial observability:
\begin{equation}
\text{VoI} = V_{\text{MDP}}(b_0) - V_{\text{POMDP}}(b_0) \geq 0.
\label{eq:voi}
\end{equation}
We note that this quantity represents the maximum additional expected reward obtainable by upgrading the classification subsystem to perfect accuracy.

\subsection{Model-Free RL Baselines}
\label{sec:rl}

To assess whether the model-based PBVI solution can be recovered without explicit use of the transition and observation models in the learning updates, we implement two model-free RL algorithms that learn policies from interaction with the POMDP environment.

\subsubsection{Q-Learning on Discretized Belief Space}
\label{sec:qlearning}

Since model-free methods cannot operate directly on a continuous belief simplex, we discretize $\Delta^{K-1}$ into a uniform grid with spacing $\delta$, producing $\binom{K - 1 + 1/\delta}{K-1}$ grid points. Each continuous belief is mapped to its nearest grid point under the $\ell_1$ norm. A standard tabular Q-learning update \cite{watkins1992qlearning} is applied:
\begin{equation}
Q(g, a) \leftarrow Q(g, a) + \alpha_{\text{ql}}\left[r + \gamma \max_{a'} Q(g', a') - Q(g, a)\right],
\label{eq:qlearning}
\end{equation}
where $g$ and $g'$ are the grid indices of the current and next belief, $r$ is the immediate reward, and $\alpha_{\text{ql}}$ is the learning rate. Exploration follows an $\varepsilon$-greedy schedule with exponential decay. The specific hyperparameters are reported in Section~\ref{sec:case_study} for the specific case study.

\subsubsection{REINFORCE Policy Gradient}
\label{sec:reinforce}

To avoid the limitations of belief discretization, we also implement the REINFORCE algorithm \cite{williams1992reinforce} with a continuous feature representation. The belief $b \in \Delta^{K-1}$ is mapped to an 11-dimensional feature vector $\phi(b)$ consisting of the 4 raw belief entries, their $\binom{4}{2} = 6$ pairwise products $b_i b_j$ ($i < j$), and the belief entropy $H(b) = -\sum_k b_k \ln(b_k + \epsilon)$. A linear softmax policy is parameterized as:
\begin{equation}
\pi_\theta(a \mid b) = \frac{\exp(\theta_a^\top \phi(b))}{\sum_{a'} \exp(\theta_{a'}^\top \phi(b))},
\label{eq:reinforce_policy}
\end{equation}
where $\theta \in \mathbb{R}^{M \times \dim(\phi)}$. Parameters are updated via the policy gradient with a running-mean baseline for variance reduction. The specific training hyperparameters are reported in Section~\ref{sec:case_study}.

Both model-free agents interact with the same Gillespie POMDP environment used by PBVI: the true state evolves according to $A^{(\pi(b))}$ where the action is selected from the agent's belief-based policy; observations are drawn from $O$; and beliefs are updated via Eq.~\eqref{eq:belief_update}. At evaluation time, both agents act greedily (Q-learning: $\arg\max_a Q(g, a)$; REINFORCE: $\arg\max_a \pi_\theta(a \mid b)$). At corner beliefs, where the true state is known with certainty, both model-free methods recover actions consistent with the MDP optimal policy, confirming that the learned policies have correctly internalized the reward structure from interaction alone.

\section{Results}
\label{sec:results}

\subsection{Case Study: Error Propagation in an ARX Digital Twin}
\label{sec:case_study}

We use the same digital twin model as in our previous study \cite{najafi2026hmm}. This modular digital twin is made up of connected surrogate modules, each represented by an ARX model. During operation, differences between the actual system outputs and those predicted by the ARX model appear as module-level residuals. The statistical patterns of these residuals change as the system moves through different hidden error states. The inference layer, as we described in the related paper \cite{najafi2026hmm}, uses a Gaussian-emission HMM to analyze these residual streams. It assigns a discrete regime label at each time step. This leads to a natural decision-making problem: given the current error regime, what corrective action, if any, should be taken to bring the system back to normal operation while minimizing unnecessary maintenance costs? We conceptualize this sequential decision problem as an MDP, where error regimes serve as states, corrective interventions serve as actions, and a scalar reward encodes the cost-benefit tradeoff between system fidelity and intervention expense.

A four-state Gaussian-emission HMM is fit to ARX-residual features from the six-module digital twin using the \texttt{depmixS4} package \cite{visser2010depmixs4} over $N_{\text{seed}} = 10$ Monte Carlo seeds, and its Viterbi decoding is Hungarian-aligned to the canonical regime labeling $\mathcal{S} = \{\textsc{Nominal}, \textsc{SensorNoisy}, \textsc{DynamicsOff}, \textsc{Drift}\}$. Per-regime classification accuracies (diagonal of the confusion matrix) are approximately $\{0.995,\, 0.21,\, 0.80,\, 0.996\}$ for \textsc{Nominal}, \textsc{SensorNoisy}, \textsc{DynamicsOff}, and \textsc{Drift} respectively, with approximately $79\%$ of \textsc{SensorNoisy} frames misclassified as \textsc{Nominal} (we note that these per-regime accuracies are averages across the 10 Monte Carlo HMM fits, weighted equally by seed. Single-seed accuracy values can differ from the aggregate). This asymmetric accuracy directly populates the observation matrix $O$ of Section~\ref{sec:observation_model} and with the EM-learned transition matrix of Section~\ref{sec:transition_model}, it closes a fully data-driven parameterization of the decision problem.

\subsubsection{State Space ($K = 4$)}
\label{sec:state_space}

The state space consists of $K = 4$ error regimes identified by the HMM inference layer:
\begin{equation}
\mathcal{S} = \{\textsc{Nominal},\; \textsc{SensorNoisy},\; \textsc{DynamicsOff},\; \textsc{Drift}\}.
\label{eq:states}
\end{equation}
\textsc{Nominal} indicates the system's operation within its specified parameters. \textsc{SensorNoisy} means there is increased measurement noise in one or more sensor channels. \textsc{DynamicsOff} indicates a change in gain or parameters within the plant dynamics module. \textsc{Drift} represents a slowly accumulating bias in a sensor output \cite{najafi2026hmm}.

\subsubsection{Action Space ($M = 6$)}
\label{sec:action_space}

The action space comprises $M = 6$ discrete corrective interventions:
\begin{multline}
\mathcal{A} = \{\textsc{NoAction},\; \textsc{DwSensorA},\; \textsc{DwSensorB},\\
\textsc{IncFilter},\; \textsc{ReidentPlant},\; \textsc{BiasCorrect}\}.
\label{eq:actions}
\end{multline}
Each action targets a specific fault mechanism within the digital twin's six-module pipeline (Actuator $\to$ Plant $\to$ Sensors $\to$ Fusion $\to$ KPI). Recall from \cite{najafi2026hmm} that the fusion layer computes $\hat{y}_{\text{fused}} = \alpha \hat{y}_A + (1 - \alpha) \hat{y}_B$ with blending coefficient $\alpha = 0.6$, and that each module has an ARX surrogate whose residuals $\epsilon_i(t) = y_{\text{ARX},i}(t) - y_{\text{Phys},i}(t)$ feed the HMM.

\textsc{NoAction} applies no intervention and is appropriate when the system is in the \textsc{Nominal} regime and residuals are within expected bounds. \textsc{DwSensorA} and \textsc{DwSensorB} reduce the contribution of sensor A (or B) in the fusion layer by decreasing its blending coefficient $\alpha$ (or $1 - \alpha$) thereby attenuating the propagation of sensor noise to downstream modules when \textsc{SensorNoisy} is active. \textsc{IncFilter} increases the smoothing bandwidth of the sensor low-pass filters (i.e., increases the filter time constant $\tau$) to attenuate high-frequency noise before the signals enter the fusion layer; this is a softer mitigation that preserves both sensor channels but sacrifices bandwidth which makes it less effective but also less disruptive when noise is moderate. \textsc{ReidentPlant} re-identifies the ARX parameters of the plant dynamics module from a recent data window, and restores fidelity between the ARX surrogate and the true plant when \textsc{DynamicsOff} is active. \textsc{BiasCorrect} subtracts the posterior-weighted bias estimate $\hat{b}(t) = \sum_k \gamma_k(t) \bar{\epsilon}_{j,k}$ from the affected sensor channel thereby removing the systematic offset when \textsc{Drift} is active.

\subsubsection{Baseline Transition Matrix (Data-Driven)}

The baseline transition matrix $A^{(\text{NoAction})}$ is calibrated from the HMM inference layer. Specifically, the EM-learned transition matrices from a 10-seed Monte Carlo simulation are extracted via the \texttt{depmixS4} package \cite{visser2010depmixs4, najafi2026hmm}, reordered to the canonical regime labeling using the Hungarian algorithm \cite{kuhn1955hungarian}, quality-filtered (retaining only seeds whose overall mapping accuracy exceeds 55\% and whose reordered transition diagonal entries all exceed $0.1$), averaged across retained seeds, and Laplace-smoothed with $\varepsilon = 10^{-3}$ to preclude absorbing states. The resulting matrix is:
\begin{equation}
A^{(\text{NoAction})} = \begin{pmatrix}
0.9970 & 0.0010 & 0.0010 & 0.0010 \\
0.0030 & 0.9846 & 0.0010 & 0.0114 \\
0.0010 & 0.0021 & 0.9071 & 0.0898 \\
0.0070 & 0.0161 & 0.1207 & 0.8562
\end{pmatrix},
\label{eq:baseline}
\end{equation}
where rows and columns are ordered as (\textsc{Nominal}, \textsc{SensorNoisy}, \textsc{DynamicsOff}, \textsc{Drift}). The diagonals (ranging from $0.856$ for \textsc{Drift} to $0.997$ for \textsc{Nominal}) reflect the persistence of each regime over the $\Delta t = 0.02$\,s sampling interval.

\subsubsection{Repair Probabilities and Reward Matrix}

Each corrective action $a$ is associated with a repair probability $\rho_{a,s} \in [0,1]$ and represents the likelihood that applying action $a$ while the system is in regime $s$ successfully transitions the system to \textsc{Nominal} in the next step. Table~\ref{tab:repair} lists the nonzero repair probabilities. These values are specified based on the physical mechanism of each intervention: sensor down-weighting has moderate efficacy because it attenuates but does not eliminate noise; filter bandwidth increase is softer still; plant re-identification is highly effective because it directly re-estimates the drifted parameters; and bias correction is the most effective because the HMM posterior provides a precise bias estimate.

\begin{table}[t]
\centering
\caption{Repair probabilities $\rho_{a,s}$ for each action-regime pair. Entries not listed are zero.}
\label{tab:repair}
\begin{tabular}{@{}llll@{}}
\toprule
Action $a$ & Mechanism & Target regime $s$ & $\rho_{a,s}$ \\
\midrule
\textsc{DwSensorA}   & Reduce $\alpha$ in fusion         & \textsc{SensorNoisy} & 0.60 \\
\textsc{DwSensorB}   & Reduce $(1 - \alpha)$ in fusion   & \textsc{SensorNoisy} & 0.60 \\
\textsc{IncFilter}   & Increase filter $\tau$             & \textsc{SensorNoisy} & 0.40 \\
\textsc{ReidentPlant}& Re-estimate $\theta_{\text{plant}}$ & \textsc{DynamicsOff} & 0.70 \\
\textsc{BiasCorrect} & Subtract $\hat{b}(t)$             & \textsc{Drift}       & 0.80 \\
\bottomrule
\end{tabular}
\end{table}

The reward function $R(a, s)$ assigns a scalar payoff for executing action $a$ in state $s$. Table~\ref{tab:reward} presents the $6 \times 4$ reward matrix.

\begin{table}[t]
\centering
\caption{Reward matrix $R(a, s)$. Positive entries indicate beneficial action-state pairings; negative entries penalize unnecessary or mismatched interventions.}
\label{tab:reward}
\begin{tabular}{@{}lcccc@{}}
\toprule
& \textsc{Nom.} & \textsc{SensNsy.} & \textsc{DynOff} & \textsc{Drift} \\
\midrule
\textsc{NoAction}      & $+1.0$ & $-0.5$ & $-1.0$ & $-0.8$ \\
\textsc{DwSensorA}     & $-0.1$ & $+0.7$ & $-0.2$ & $-0.1$ \\
\textsc{DwSensorB}     & $-0.1$ & $+0.6$ & $-0.2$ & $-0.1$ \\
\textsc{IncFilter}     & $-0.1$ & $+0.4$ & $ 0.0$ & $+0.1$ \\
\textsc{ReidentPlant}  & $-0.5$ & $-0.3$ & $+0.9$ & $-0.2$ \\
\textsc{BiasCorrect}   & $-0.2$ & $ 0.0$ & $+0.1$ & $+0.8$ \\
\bottomrule
\end{tabular}
\end{table}

We design the structure of the reward matrix in order to reflect three design principles. First, the highest reward in the \textsc{Nominal} state is achieved by \textsc{NoAction} ($+1.0$) and discourages unnecessary interventions when the system is healthy. Second, each fault state awards the largest reward to its targeted corrective action (e.g., \textsc{DwSensorA} yields $+0.7$ in \textsc{SensorNoisy}, \textsc{ReidentPlant} yields $+0.9$ in \textsc{DynamicsOff}, and \textsc{BiasCorrect} yields $+0.8$ in \textsc{Drift}). Third, applying an unrelated or costly action in a fault state incurs a small negative penalty thereby reflecting wasted computational resources or unnecessary model perturbation.

\subsubsection{Observation Model (Data-Driven)}

The observation matrix $O \in \mathbb{R}^{K \times K}$ with $O_{s,z} = P(\text{observe } z \mid \text{true state } s)$ is derived from the Viterbi confusion matrix of the HMM decoder evaluated against ground-truth regime labels across all Monte Carlo seeds. Each row sums to unity. The \textsc{DynamicsOff} regime is classified with approximately 80\% accuracy and \textsc{Drift} with approximately 99.6\% accuracy, because both induce distinct residual signatures (mean shifts in gain and bias, respectively). In contrast, the \textsc{SensorNoisy} regime achieves only approximately 21\% classification accuracy because elevated sensor variance produces a variance increase without a large mean shift in the residual features; roughly 79\% of \textsc{SensorNoisy} observations are misclassified as \textsc{Nominal}. This asymmetric observation quality is a consequence of the physical signal characteristics, and it is the primary driver of the performance gap between MDP and POMDP. In our setting, this gap is driven primarily by the low classification accuracy for \textsc{SensorNoisy}, which creates persistent belief uncertainty and delays appropriate sensor-repair actions. The \textsc{DynamicsOff} and \textsc{Drift} regimes, classified accurately enough that their transitions are detected within one or two observations, contribute minimally to the observability cost.

\subsubsection{Solver Parameters}

We set the discount factor to $\gamma = 0.99$ resulting in an effective planning horizon of $1/(1 - \gamma) = 100$ discrete steps, or equivalently 2\,s at the $\Delta t = 0.02$\,s sampling rate. With $\gamma = 0.99$ and $\Delta t = 0.02$\,s, we obtain $\rho = -\ln(0.99)/0.02 \approx 0.503$\,s$^{-1}$. Value iteration converges in approximately 2,300 iterations with tolerance $10^{-10}$.

The resulting optimal policy is intuitive where we perform \textsc{NoAction} in \textsc{Nominal}, \textsc{DwSensorA} in \textsc{SensorNoisy}, \textsc{ReidentPlant} in \textsc{DynamicsOff}, and \textsc{BiasCorrect} in \textsc{Drift}. The state values are nearly uniform ($V^* \approx 99.5$ for all states), which we attribute to the high repair probabilities that allow rapid return to \textsc{Nominal} from any fault state.

For PBVI, we generate approximately 200 belief points by combining $K = 4$ corner beliefs (pure states), $\binom{K}{2} = 6$ edge midpoints, $\binom{K}{3} = 4$ face centroids, the simplex centroid $(1/K, \ldots, 1/K)$, and approximately 185 random Dirichlet$(1, \ldots, 1)$ samples. PBVI is run for a fixed budget of 500 Bellman iterations and retains approximately 108 unique $\alpha$-vectors after pruning, and the resulting policy matches the MDP-optimal action at all four corner beliefs. Figure~\ref{fig:pbvi_convergence} reports the full convergence trace, $\delta_k = \max_{b \in \mathcal{B}}\, |V_k(b) - V_{k-1}(b)|$. The MDP warm start drives $\delta_k$ from $\delta_0 \approx 0.9$ down to $\approx 10^{-1}$ within roughly $20$ iteration, after which $\delta_k$ plateaus in the $10^{-2}$ band and the run terminates at the iteration cap $k = 500$. The point-based Bellman operator is non-contracting on the fixed belief set $\mathcal{B}$ and the $\alpha$-vector set is updated asymmetrically as belief points enter and leave the support of different actions, so the residual need not collapse to machine precision even though the greedy policy stabilizes. We therefore report convergence by policy stability rather than a residual tolerance. The greedy policy extracted from the $\alpha$-vector set reaches its final mapping within the first few dozen iterations and remains unchanged for the remainder of the budget, which is the criterion used throughout Section~\ref{sec:results}.

For Q-learning, we use $\delta = 0.1$ (producing $\binom{K - 1 + 1/\delta}{K-1} = 286$ grid points for $K = 4$), learning rate $\alpha_{\text{ql}} = 0.1$, and $\varepsilon$-greedy exploration decaying from $\varepsilon = 1.0$ to $\varepsilon = 0.05$ over 5,000 episodes each of duration $T = 15$\,s. For REINFORCE, the belief is mapped to an 11-dimensional feature vector $\phi(b)$ consisting of the 4 raw belief entries, their $\binom{4}{2} = 6$ pairwise products $b_i b_j$ ($i < j$), and the belief entropy $H(b) = -\sum_k b_k \ln(b_k + \epsilon)$. Training runs for 10,000 episodes with learning rate $\alpha_{\text{re}} = 0.001$ and running-mean baseline decay 0.99.

\revnew{\begin{figure}[htbp]
\centering
\includegraphics[width=\columnwidth]{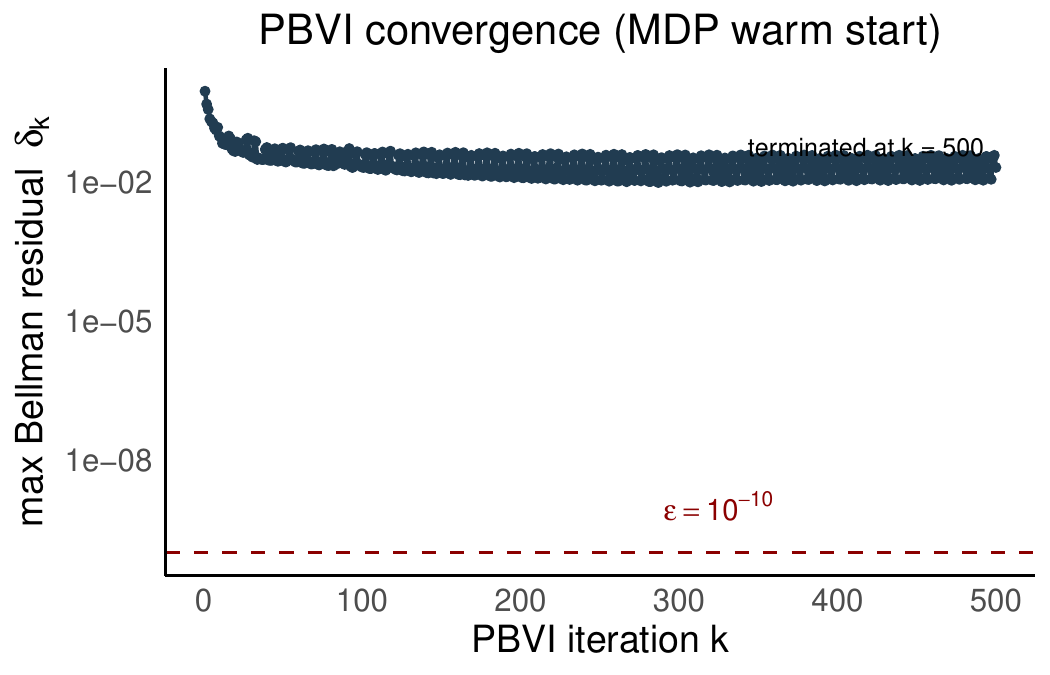}
\caption{PBVI convergence trajectory under the MDP warm start (Section~\ref{sec:pbvi}, case-study observation matrix). The curve plots the maximum Bellman residual $\delta_k = \max_{b \in \mathcal{B}} |V_k(b) - V_{k-1}(b)|$ on a log scale against iteration $k$. The residual drops below $10^{-1}$ within a handful of iterations under the MDP warm start and then oscillates in the $10^{-2}$ band because the point-based operator is non-contracting on the fixed belief grid. The run terminates at the iteration cap $k = 500$ (annotated on the figure) rather than by residual-based convergence. The $\varepsilon = 10^{-10}$ reference line indicates the order of magnitude at which value iteration on the MDP does converge (Section~\ref{sec:value_iteration}) and is never reached by PBVI here. The greedy policy, however, stabilizes well inside the $500$-iteration budget.}
\label{fig:pbvi_convergence}
\end{figure}}

\subsubsection{MCDA/TOPSIS Baseline}
\label{sec:mcda_baseline}

The MCDA baseline uses the TOPSIS action-selection component of the companion paper's intervention pipeline \cite{najafi2026hmm} and is implemented using the \texttt{apply.TOPSIS} function from the \texttt{RMCDA} package \cite{najafi2025rmcda}.  The companion paper's pipeline comprises two stages: posterior-weighted bias correction subtracted from surrogate outputs at the module level, and TOPSIS ranking over discrete corrective actions. We note that only the second stage is relevant as a baseline here because the MDP, POMDP, and RL agents all select actions from a shared discrete menu without applying module-level bias correction. Comparing against MCDA's full pipeline would therefore mix model-level correction with action selection and conflate the two sources of performance improvement. At each transition event, an exponential moving average (smoothing factor $0.3$) over one-hot observation indicators produces a posterior estimate $\hat{p}$ over the four regimes. The five corrective actions \{\textsc{DwSensorA}, \textsc{DwSensorB}, \textsc{IncFilter}, \textsc{ReidentPlant}, \textsc{BiasCorrect}\} are scored on five criteria (KPI gain, Module gain, Drift gain, Stability, and Cost) computed from $\hat{p}$, with the same functional form and weights ($0.35, 0.25, 0.15, 0.10, 0.15$) used in \cite{najafi2026hmm}. \texttt{RMCDA::apply.TOPSIS} then ranks the five actions and the top-ranked action is selected. When the posterior on \textsc{Nominal} exceeds $0.75$ and the KPI proxy is below $0.25$, the action is overridden to \textsc{NoAction}, matching the idle gate of the companion paper. A time-based dwell hysteresis of $\tau_{\text{dwell}} = 0.5$\,s prevents rapid action switching.

\subsubsection{$k$-Step Consecutive Heuristic Baseline}
\label{sec:kstep_baseline}

The \textsc{NoAction}-always policy corresponds to a controller that ignores the HMM layer entirely, and arguably understates what a reasonable operator would do without access to an optimization framework. To obtain a stricter responsive baseline we add a $k$-step consecutive-non-\textsc{Nominal} heuristic where the operator receives the same noisy HMM label stream as the MCDA and POMDP controllers, applies \textsc{NoAction} while the stream is \textsc{Nominal} or oscillating, and commits to the canonical repair for an observed regime $z \neq \textsc{Nominal}$ only after $z$ has been reported for $k$ consecutive transition events. A single \textsc{Nominal} observation resets the counter, and the same $\tau_{\text{dwell}} = 0.5$\,s hysteresis of Section~\ref{sec:mcda_baseline} prevents flip-flopping. We sweep $k \in \{1, 2, 3, 5\}$ and select the value that maximizes mean discounted return on a held-out set of 1{,}000 seeds. The canonical repairs are \textsc{DwSensorA} for \textsc{SensorNoisy}, \textsc{ReidentPlant} for \textsc{DynamicsOff}, and \textsc{BiasCorrect} for \textsc{Drift}, mirroring the MDP-optimal policy at the corner beliefs.

\begin{figure*}[htbp]
\centering
\includegraphics[width=0.75\textwidth]{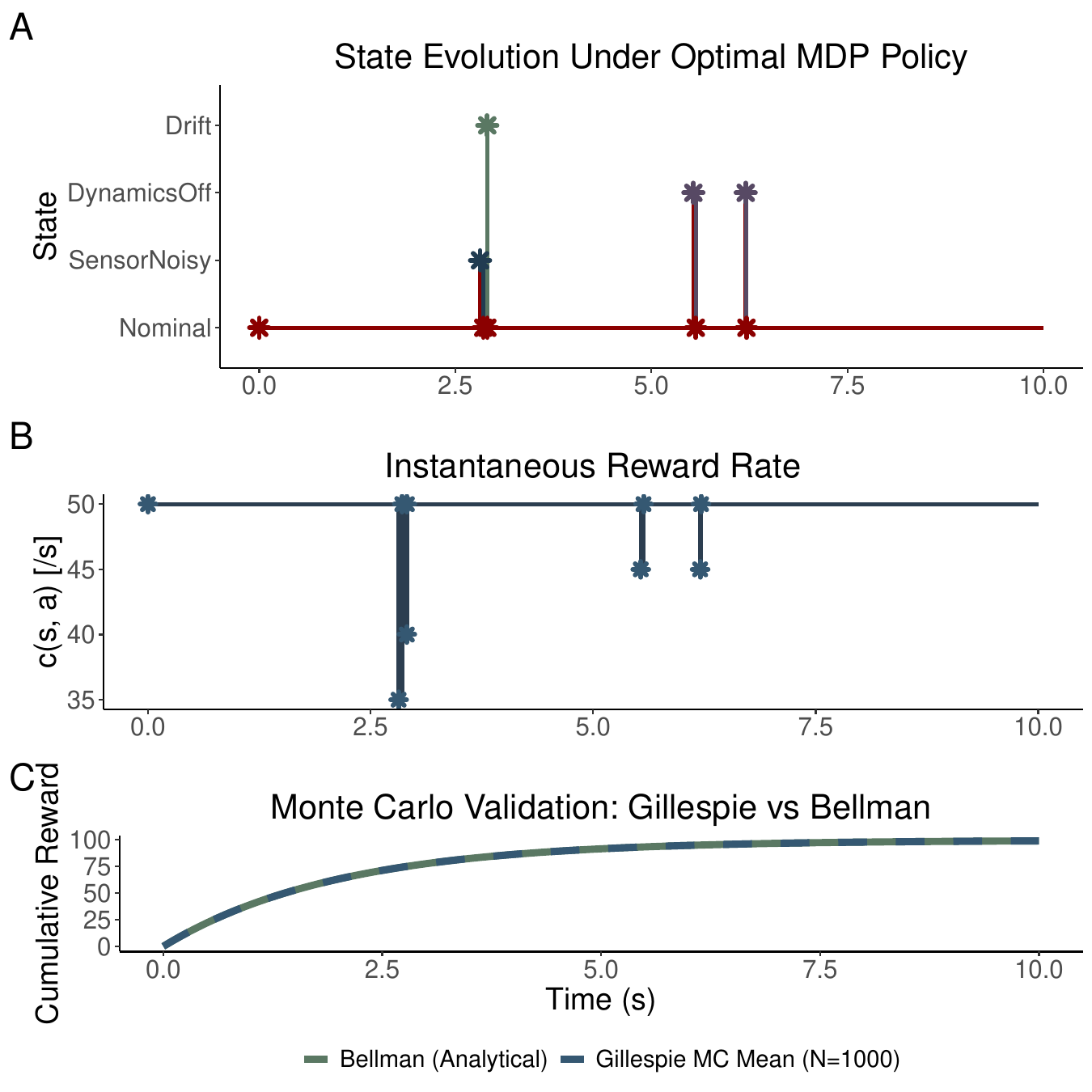}
\caption{MDP validation and trajectory analysis. (A) state evolution under the optimal policy (seed 13), showing rapid fault recovery. (B) instantaneous reward rate, confirming correct action selection in each regime. (C) Monte Carlo validation (1,000 trajectories) against the closed-form Bellman solution, with relative error below 0.02\%.}
\label{fig:trajectory}
\end{figure*}

We present the foundational validation of the MDP framework in Figure~\ref{fig:trajectory}, which confirm the correctness and consistency of our implementation through three independent computational paths. In Figure~\ref{fig:trajectory}A we show the state evolution of a representative 10-second Gillespie trajectory under the optimal MDP policy. The system starts in \textsc{Nominal} and occasionally transitions to fault states (\textsc{SensorNoisy}, \textsc{DynamicsOff}, or \textsc{Drift}), at which point the optimal policy immediately prescribes the correct corrective action. Each fault excursion is brief (typically less than 0.5\,s) because the repair probabilities are high (0.60 to 0.80), giving each transition a high chance of returning the system to \textsc{Nominal}.

Figure~\ref{fig:trajectory}B displays the instantaneous reward rate $c(s, \pi^*(s))$ over the same trajectory. In \textsc{Nominal} with \textsc{NoAction}, the rate is $+50$/s (the maximum possible at $R = +1.0$ divided by $\Delta t = 0.02$\,s). When a fault occurs, the rate drops to the reward rate of the correct repair action in that fault state (e.g., $+35$/s for \textsc{DwSensorA} in \textsc{SensorNoisy}). The sharp drops and rapid recoveries correspond exactly to the fault excursions visible in Panel~A, confirming that the policy responds correctly to every regime transition.

In Figure~\ref{fig:trajectory}C, we ran 1,000 independent Gillespie trajectories starting from \textsc{Nominal} and computed the Monte Carlo mean of the cumulative discounted reward at each evaluation point (solid line). We overlaid the closed-form transient Bellman solution computed via the matrix exponential from Eq.~\eqref{eq:theory_curve} (dashed line). The simulations and theoretical forms were indistinguishable, with relative error below 0.02\% at $t = 10$\,s. This agreement between the stochastic Gillespie simulator and the analytical Bellman solution, where both consume the same first-order CTMC generator $Q^{(a)}$ of Section~\ref{sec:ctmc} confirms numerical consistency. We note that the match certifies implementation consistency within the continuous-time pipeline rather than exact fidelity to the original discrete-time MDP (a separate comparison against discrete-time value iteration, which also shows sub-percent agreement, is reported alongside the code release). We treat this combined internal-consistency evidence as sufficient grounding to build subsequent analyses on.

\begin{figure*}[htbp]
\centering
\includegraphics[width=0.75\textwidth]{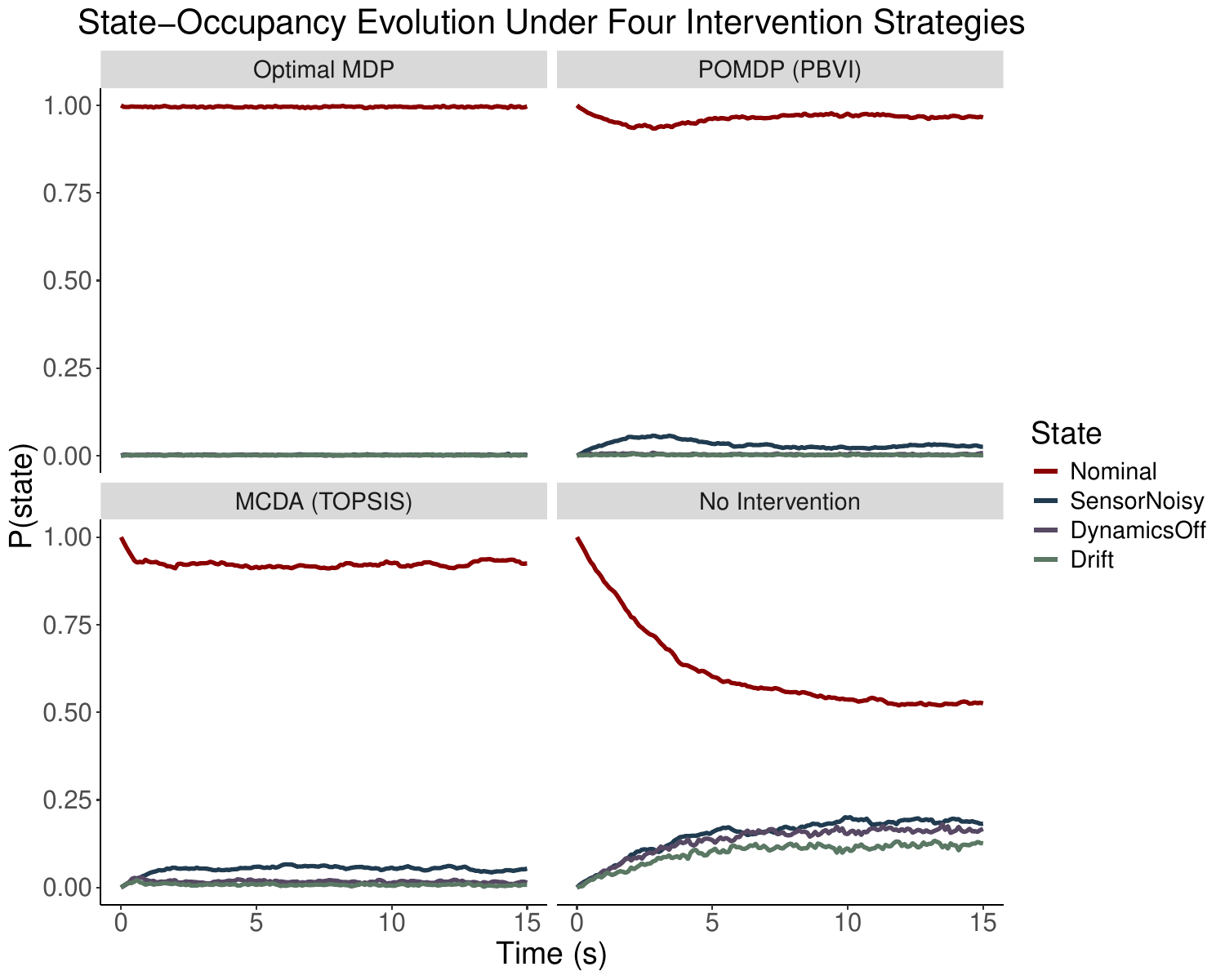}
\caption{State-occupancy evolution under four intervention strategies (2,000 trajectories each, starting from \textsc{Nominal}). The $2 \times 2$ layout reveals the progressive degradation from perfect observability (MDP) through partial observability (POMDP) to hard classification (MCDA) and inaction.}
\label{fig:state_occupancy}
\end{figure*}

To understand how each policy shapes the long-run statistical behavior of the system, we simulated 2,000 independent Gillespie trajectories per policy over 15\,s and tracked the empirical state-occupation probability $P(\text{state} = s \mid \text{time} = t)$ at each time bin. Figure~\ref{fig:state_occupancy} displays these probabilities in a $2 \times 2$ grid of panels, one per policy: Optimal MDP, POMDP (PBVI), MCDA (TOPSIS), and No Intervention.

Under the Optimal MDP, the \textsc{Nominal} probability stays above 0.95 throughout the entire 15-second window. The three fault states have very low probabilities (each below 0.02), indicating that the system spends nearly all its time in the healthy regime. The slight drop in \textsc{Nominal} probability at early times is reflective of the transient phase where the system has not yet reached its stationary distribution; by $t \approx 2$\,s, the probabilities converge to their steady-state values. This near-perfect \textsc{Nominal} occupancy is reflective of the combination of high repair probabilities and immediate, always-correct action selection.

Under the POMDP model, the system maintains high \textsc{Nominal} occupancy but with a visible gap relative to the MDP. The \textsc{Nominal} probability settles to approximately 0.90-0.93, and the fault-state probabilities are correspondingly higher. This gap represents the cost of partial observability: because the POMDP agent perceives the system through noisy classifications rather than the true state, it sometimes delays or misidentifies corrective actions, particularly for \textsc{SensorNoisy}. Despite this, the POMDP's Bayesian belief maintenance enables it to recover most of the MDP's performance.

Under MCDA, the system has noticeably lower \textsc{Nominal} occupancy. The state probabilities are more spread across regimes because MCDA frequently misclassifies the regime and applies wrong actions. However, an important subtlety emerges: MCDA's \textsc{Nominal} occupancy is not as catastrophically low as its reward performance might suggest. This is because mismatched actions applied while in \textsc{Nominal} do not change the transition dynamics, since repair actions only have effect in their target fault state, so the system can remain in \textsc{Nominal} even when the wrong action is applied. The reward penalty from these mismatched actions is severe, earning $-0.1/\Delta t$ to $-0.5/\Delta t$ instead of $+1.0/\Delta t$, but the physical state is unaffected.

Under no intervention, the system evolves according to the baseline transition matrix $A^{(\text{NoAction})}$ with no corrective actions. Starting from \textsc{Nominal}, the system gradually diffuses across all four regimes and eventually approaches the stationary distribution of the baseline chain. By $t = 15$\,s, the system occupies each state with a non-uniform stationary distribution (${\approx}$52\% \textsc{Nominal}, 20\% \textsc{SensorNoisy}, 16\% \textsc{DynamicsOff}, 11\% \textsc{Drift}) which reflects the asymmetric outflow structure of the baseline transition matrix.

Next, we conducted the central performance comparison by simulating 1,000 Gillespie trajectories per policy over a 20-second horizon under noisy observations governed by the confusion matrix.

\begin{figure*}[htbp]
\centering
\includegraphics[width=0.85\textwidth]{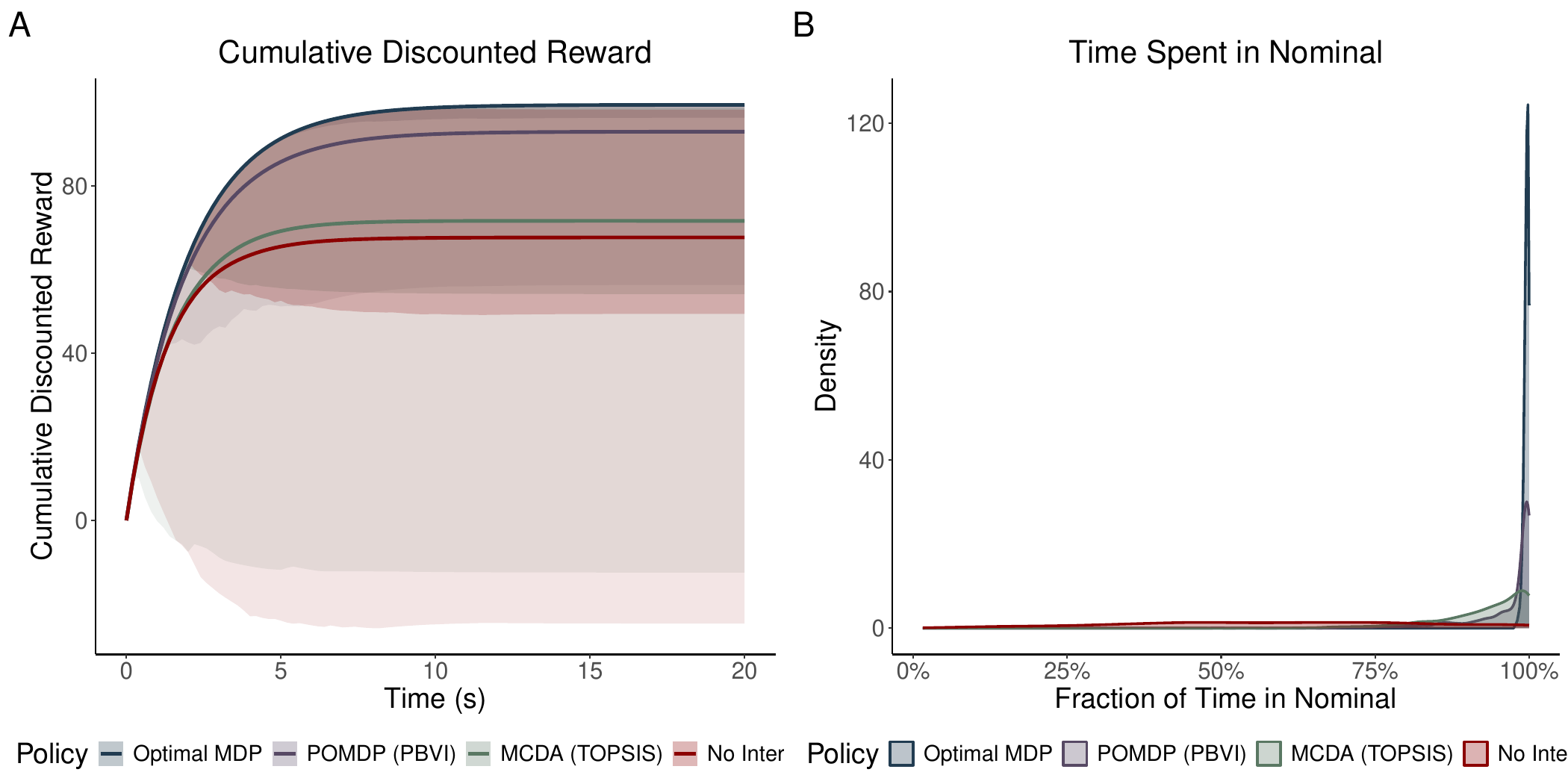}
\caption{Policy performance comparison across 1,000 trajectories. (A) cumulative discounted reward over time (dark band = IQR, light band = 5th-95th percentile). (B) distribution of fraction of time spent in \textsc{Nominal} (dashed lines = means). The POMDP recovers most of the MDP's advantage; MCDA collapses to near No Intervention.}
\label{fig:reward_comparison}
\end{figure*}

In Figure~\ref{fig:reward_comparison}A we show the mean cumulative discounted reward over time for four policies, with shaded bands showing the interquartile range. The Optimal MDP achieves the highest reward with the tightest confidence bands which shows the near-perfect outcome under perfect state knowledge. The POMDP tracks close behind, recovering approximately 95\% of MDP performance which is a remarkable result given the 21\% \textsc{SensorNoisy} classification accuracy. The MCDA and No Intervention curves cluster together at dramatically lower levels, with MCDA barely outperforms complete inaction. This collapse occurs because MCDA's hard classification under the data-driven confusion matrix produces interventions that are wrong more often than they are right. As a result the negative rewards from mismatched actions nearly cancel the positive rewards from occasional correct ones. However, as we explain next, it does not necessarily indicate that the system is healthier with no intervention compared to the MCDA performance.

In Figure~\ref{fig:reward_comparison}B, we track the fraction of total simulation time that each policy keeps the system in the \textsc{Nominal} state, and show the results as kernel density plots with dashed vertical lines indicating mean values. Unlike cumulative reward, this is an undiscounted quantity that directly measures system fidelity independent of the reward structure. The MDP distribution is a sharp spike near $99.5\%$, indicating near-deterministic healthy operation. The POMDP distribution is broader, centered around $93\%$ to $96\%$, with a left tail reflecting occasional extended fault episodes when the belief filter is slow to detect \textsc{SensorNoisy}. The MCDA and No Intervention distributions are broader and centered at substantially lower values. MCDA's \textsc{Nominal} occupancy, while lower than the POMDP's, is not catastrophically low; the system does spend a meaningful fraction of time in \textsc{Nominal}. Yet its cumulative reward is barely above No Intervention.

\begin{figure*}[htbp]
\centering
\includegraphics[width=0.85\textwidth]{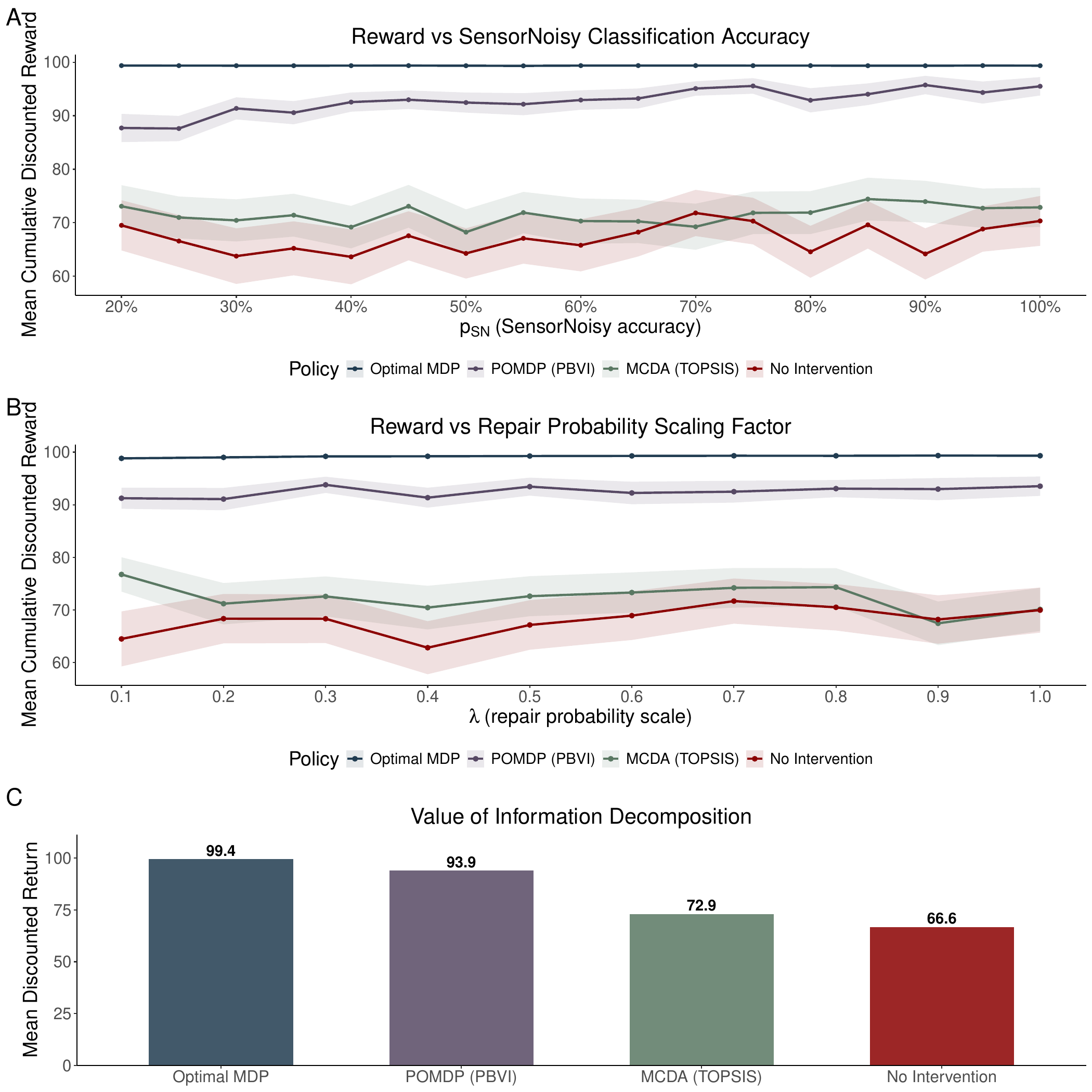}
\caption{Sensitivity analysis and value of information. (A) reward vs.\ \textsc{SensorNoisy} classification accuracy, showing the POMDP's degradation and persistent planning advantage over MCDA. (B) reward vs.\ repair probability scaling factor, confirming hierarchy robustness. (C) value of information decomposition quantifying the cost of partial observability (${\approx}$5.5) vs.\ the planning advantage (${\approx}$21).}
\label{fig:sensitivity}
\end{figure*}

We then continued by conducting three complementary analyses to stress-test the policy hierarchy and decompose the sources of performance differences. First, we swept the \textsc{SensorNoisy} classification accuracy $p_{\text{SN}}$ from 20\% to 100\% in steps of 5\%, modifying the \textsc{SensorNoisy} row of the observation matrix at each level while holding all other rows at their data-driven values. At each accuracy level, we re-solved the POMDP via PBVI (since the observation model changed) and simulated 300 trajectories per policy. The MDP (Figure~\ref{fig:sensitivity}A) is invariant to observation quality because it knows the true state. The POMDP rises monotonically from lower values at $p_{\text{SN}} = 20\%$ toward the MDP bound as accuracy approaches 100\% which demonstrates degradation with observation quality. MCDA also improves with better observations but starts from a dramatically lower baseline and remains below the POMDP at every accuracy level which confirms that even with perfect sensors, MCDA's myopic hard classification cannot match the POMDP's belief-based planning. The vertical gap between the POMDP and MCDA curves quantifies the planning advantage, the value of maintaining soft beliefs and reasoning about future uncertainty, which is present at all accuracy levels but is largest at low $p_{\text{SN}}$ where the consequences of hard classification are most severe.

We acknowledge that the repair probabilities in our formulation were specified by engineering judgment rather than estimated from maintenance records. As a result, to test robustness, we scaled all repair probabilities by a common factor $\lambda \in [0.1, 1.0]$ in steps of 0.1. For each $\lambda$, we rebuilt the action-dependent transition matrices, re-ran value iteration and PBVI, and simulated 300 trajectories per policy. The policy hierarchy (MDP $>$ POMDP $\gg$ MCDA $\approx$ No Intervention) is preserved at every $\lambda$ value (Figure~\ref{fig:sensitivity}B). The optimal policy mapping also remains stable across all $\lambda$ values (\textsc{NoAction} in \textsc{Nominal}, \textsc{DwSensorA} in \textsc{SensorNoisy}, \textsc{ReidentPlant} in \textsc{DynamicsOff}, \textsc{BiasCorrect} in \textsc{Drift}) thereby confirming that the reward structure identifies the best action regardless of repair efficacy. The approximately linear relationship between $\lambda$ and performance indicates that incremental improvements in maintenance reliability translate directly to proportional gains without diminishing returns.

Because the $R(a, s)$ entries of Table~\ref{tab:reward} are themselves engineering choices, we perform an additional sensitivity analysis with a three-way perturbation of the reward matrix itself: (i) uniform scaling $R' = \alpha R$ with $\alpha \in \{0.5,\, 1.0,\, 2.0\}$, which contracts or stretches the full reward scale; (ii) penalty-only scaling in which every negative entry is multiplied by $\beta \in \{0.5,\, 1.0,\, 2.0\}$ while positive entries are held fixed, stress-testing the balance between mismatch cost and repair benefit; and (iii) additive Gaussian perturbation $R' = R + \mathcal{N}(0, \sigma^2)$ with $\sigma \in \{0.05,\, 0.10,\, 0.20\}$ and 20 replicates per $\sigma$, probing sensitivity to unstructured noise in the reward specification. For every perturbation we re-solve value iteration and PBVI (the optimal action mapping can change under an aggressive reward deformation), regenerate the MCDA and $k$-step baselines, and simulate 200 trajectories per policy. The resulting rankings $\text{MDP} > \text{POMDP} > k\text{-Step} > \text{MCDA} > \text{NoAction}$ are mostly preserved in all three perturbation families, and the optimal MDP policy mapping is stable for $\alpha \in [0.5, 2.0]$, $\beta \in [0.5, 2.0]$, and $\sigma \leq 0.10$ (at $\sigma = 0.20$ the policy occasionally switches in states where two actions have nearly tied rewards, but the hierarchy of policy returns is retained). Figure~\ref{fig:reward_sensitivity} summarizes the three sweeps.

\revnew{\begin{figure*}[htbp]
\centering
\includegraphics[width=0.95\textwidth]{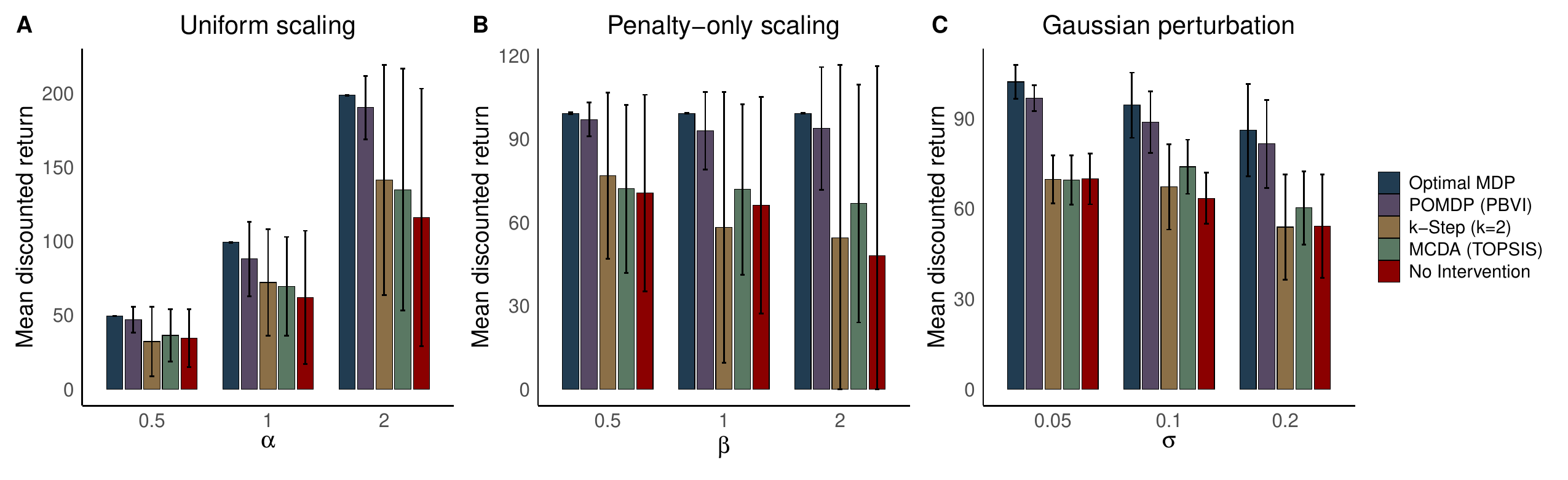}
\caption{Reward sensitivity analysis across three perturbation families. (A) uniform scaling $\alpha$, (B) penalty-only scaling $\beta$, and (C) additive Gaussian noise $\sigma$ (error bars denote the across-replicate standard deviation at each $\sigma$). The $\text{MDP} > \text{POMDP} > \text{MCDA} > \text{NoAction}$ hierarchy is mostly preserved at every level.}
\label{fig:reward_sensitivity}
\end{figure*}}

We simulated 1,000 trajectories per policy over 20\,s and computed the mean discounted return, displayed as a bar chart (Figure~\ref{fig:sensitivity}C). The bar heights show the hierarchy and quantify three distinct sources of performance loss. The MDP-POMDP gap (${\approx}$5.5 reward units) is the cost of partial observability, the maximum additional reward achievable by upgrading to perfect classification. The POMDP-MCDA gap (${\approx}$21 reward units) is the planning advantage, the value of maintaining soft beliefs rather than committing to hard classifications. The MCDA-No Intervention gap is modest (${\approx}$72.9 vs.\ ${\approx}$66.6), which demonstrates that MCDA's hard classification under this observation matrix provides only a small benefit over complete inaction (we note that changing the reward function would affect the results). The most actionable finding is that the planning advantage (${\approx}$21 units) is approximately four times larger than the observability cost (${\approx}$5.5 units).

\begin{figure*}[htbp]
\centering
\includegraphics[width=0.75\textwidth]{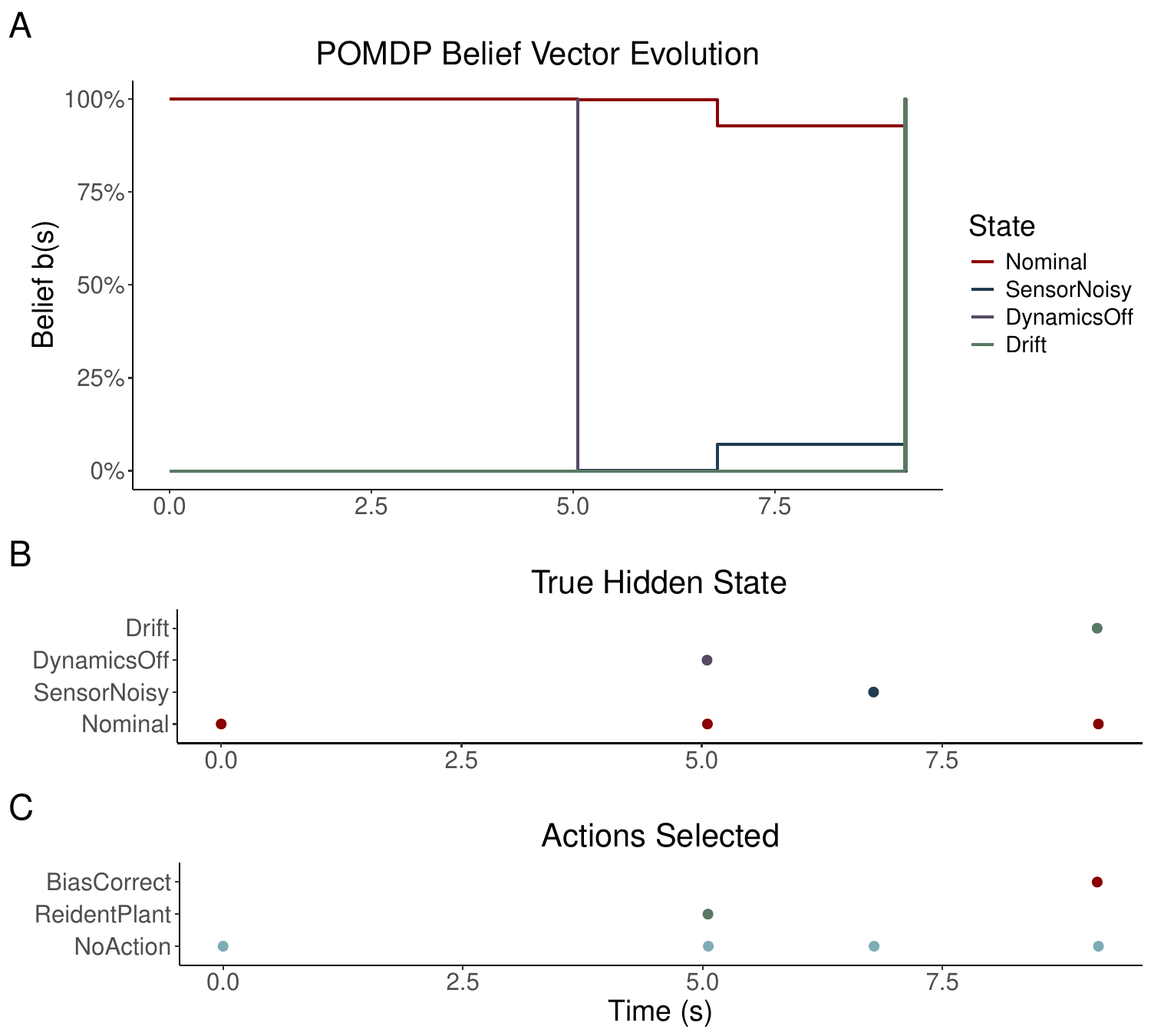}
\caption{POMDP belief trajectory visualization for a trajectory visiting all four states (selected from 500 seeds). (A) belief vector components evolving over time, showing rapid detection of \textsc{DynamicsOff}/\textsc{Drift} but slow detection of \textsc{SensorNoisy}. (B) true hidden state (ground truth). (C) actions selected by the POMDP policy.}
\label{fig:belief_trajectory}
\end{figure*}

We described the POMDP's Bayesian belief update mathematically in Section~\ref{sec:belief_update}, but we wanted to build intuition for how the belief vector actually behaves during operation and to identify the mechanistic bottleneck that prevents the POMDP from achieving MDP-level performance. Consequently, we implemented a Gillespie simulator that records the full $K$-dimensional belief vector at every transition event, alongside the true hidden state, noisy observations, and selected actions. We searched across 500 random seeds for a trajectory that visits all four states within 10\,s in order to ensure a comprehensive illustration of the belief dynamics. Figure~\ref{fig:belief_trajectory} displays the result in three synchronized panels.

The four colored step functions show the belief components $b(\textsc{Nominal})$, $b(\textsc{SensorNoisy})$, $b(\textsc{DynamicsOff})$, and $b(\textsc{Drift})$ evolving over time. At any instant, these four values sum to 1.0. When the belief is concentrated on one state (one line near 1.0, others near 0.0), the agent is confident about the current regime and acts decisively. When the belief is spread across multiple states, the agent is uncertain and acts conservatively. The most important pattern is the asymmetry in detection speed where transitions to \textsc{DynamicsOff} and \textsc{Drift} are detected within one or two observations (the corresponding belief component rises rapidly) because these regimes have high classification accuracy (80\% and 99.6\%, respectively). Transitions to \textsc{SensorNoisy}, however, produce a slow, gradual rise in $b(\textsc{SensorNoisy})$ while $b(\textsc{Nominal})$ decays which is reflective of the high confusion rate between these two regimes. The colored dots show the actual regime at each transition event, the ground truth that the agent does not observe directly. Comparing Panel~B to Panel~A reveals how well (or poorly) the belief tracks reality for each regime type. The colored dots show the action chosen by the POMDP policy at each event, based solely on the current belief vector. When the belief is confident and correct, the action matches the MDP optimal policy. When the belief is uncertain (e.g., split between \textsc{Nominal} and \textsc{SensorNoisy}), the POMDP typically selects \textsc{NoAction}, the rational choice because \textsc{NoAction} earns $+1.0$ in \textsc{Nominal} but only $-0.5$ in \textsc{SensorNoisy}, while \textsc{DwSensorA} earns $-0.1$ in \textsc{Nominal} and $+0.7$ in \textsc{SensorNoisy}. Under a 50/50 belief, the immediate expected reward slightly favors \textsc{DwSensorA} ($+0.30$ vs.\ $+0.25$ for \textsc{NoAction}). However, the POMDP policy is determined by the $\alpha$-vector value function, which incorporates discounted future consequences in addition to immediate reward. This conservative behavior is optimal given the reward structure and directly explains why the POMDP outperforms MCDA: it avoids the costly mismatched actions that plague hard classification.

The \textsc{SensorNoisy} detection lag visible in Panel~A explains the remaining MDP-POMDP gap identified in Figure~\ref{fig:sensitivity}. Improving the HMM's ability to distinguish \textsc{SensorNoisy} from \textsc{Nominal} would accelerate belief convergence and yield the largest marginal improvement in POMDP performance.

Figures~\ref{fig:reward_comparison} and~\ref{fig:sensitivity} showed that the POMDP framework dramatically outperforms MCDA, but they did not explain the mechanism. We considered two competing hypotheses:
\begin{enumerate}
    \item (H1) the POMDP intervenes more aggressively and applied corrections more frequently.
    \item (H2) the POMDP intervenes more selectively, applying fewer but better-targeted corrections.
\end{enumerate}
To distinguish between these hypotheses, we simulated 500 trajectories per policy over 20\,s and computed the action mismatch rate or the time-weighted fraction of each trajectory during which the policy applies an action that is not optimal for the current true state. We defined correct according to the MDP optimal policy (\textsc{NoAction} in \textsc{Nominal}, \textsc{DwSensorA} in \textsc{SensorNoisy}, \textsc{ReidentPlant} in \textsc{DynamicsOff}, \textsc{BiasCorrect} in \textsc{Drift}).

\begin{figure}[htbp]
\centering
\includegraphics[width=\columnwidth]{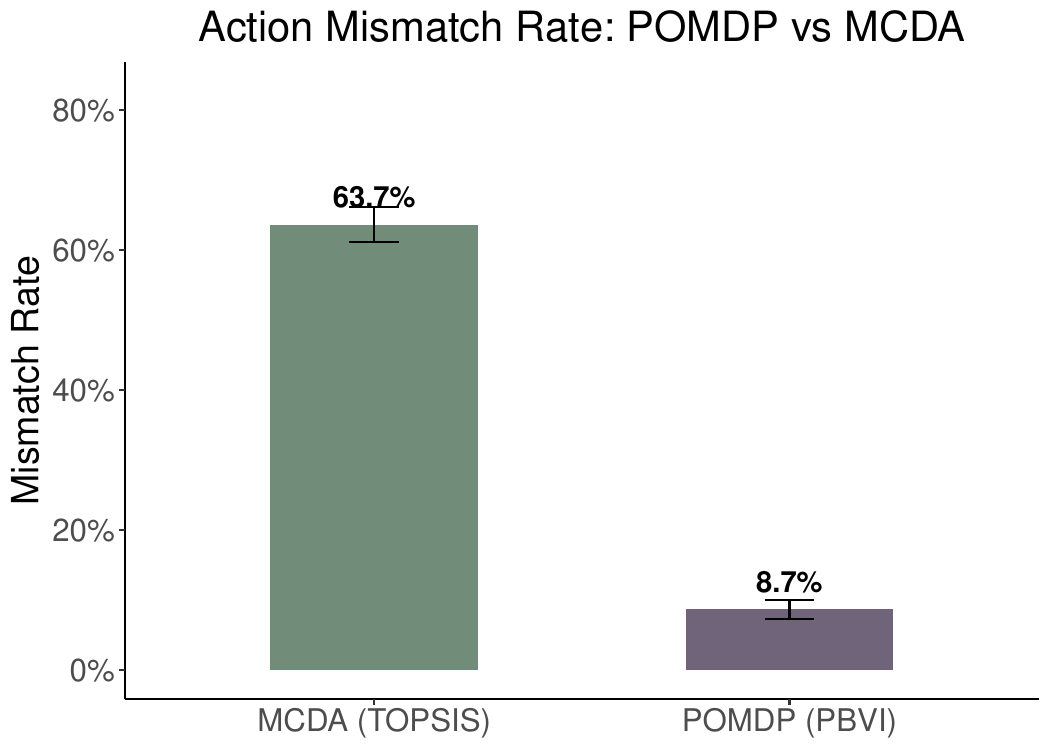}
\caption{Action mismatch rate: fraction of time each policy applies a non-optimal action (500 trajectories, 20\,s each). The POMDP achieves ${\approx}$$8.7\%$ mismatch vs.\ MCDA's ${\approx}$$63.7\%$, confirming that it wins through precision, not aggression.}
\label{fig:action_mismatch}
\end{figure}

In Figure~\ref{fig:action_mismatch}, we present the results as a bar chart with 95\% confidence intervals. The POMDP achieves a mismatch rate of approximately 8.7\%, meaning it applies the correct action over 91\% of the time. MCDA has a mismatch rate of approximately 63.7\%, applying the wrong action more often than the right one. These results strongly support H2 and the fact that the POMDP wins through precision and not aggression.

The POMDP's low mismatch arises from its conservative belief-based strategy. When the belief is concentrated on the correct state (which happens quickly for \textsc{DynamicsOff} and \textsc{Drift}), it selects the correct repair action. When the belief is uncertain (typically split between \textsc{Nominal} and \textsc{SensorNoisy}), it defaults to \textsc{NoAction}, which is the correct action whenever the system is actually in \textsc{Nominal}. Since the system spends most of its time in \textsc{Nominal} under the POMDP's effective repair policy, this conservative default is correct most of the time.

MCDA's high mismatch arises from three compounding factors. First, with only 21\% \textsc{SensorNoisy} accuracy, 79\% of \textsc{SensorNoisy} observations are misclassified as \textsc{Nominal}, leading to \textsc{NoAction} when \textsc{DwSensorA} is needed. Second, MCDA can misclassify \textsc{Nominal} as a fault state and apply unnecessary repair actions that incur negative rewards. Third, the dwell-time hysteresis ($\tau_{\text{dwell}} = 0.5$\,s) means that once MCDA locks onto a wrong action, it remains committed even as subsequent observations suggest otherwise. The $63.7\%$ mismatch rate explains MCDA's reward collapse: for roughly 13 out of every 20 seconds, the system earns suboptimal or negative rewards.

Lastly, we note that we had fixed the discount factor at $\gamma = 0.99$ throughout our analysis which corresponds to an effective planning horizon of 100 steps (2\,s). We wanted to verify that this choice did not bias our conclusions. Specifically, we hypothesized that the optimal policy and the policy hierarchy would be invariant to $\gamma$ because the reward structure and transition dynamics, not the planning horizon, determine which action is best in each state.

To test this, we swept $\gamma \in \{0.90, 0.95, 0.99, 0.999\}$, spanning effective horizons from 10 steps (0.2\,s) to 1,000 steps (20\,s). For each value, we re-ran value iteration (to check if the optimal policy changes), re-solved the POMDP via PBVI (with the new discount factor), and simulated 300 trajectories per policy. We measured the fraction of time in \textsc{Nominal}, an undiscounted metric that reflects the physical behavior of the system independent of how we account for future rewards.

\begin{figure*}[htbp]
\centering
\includegraphics[width=0.65\textwidth]{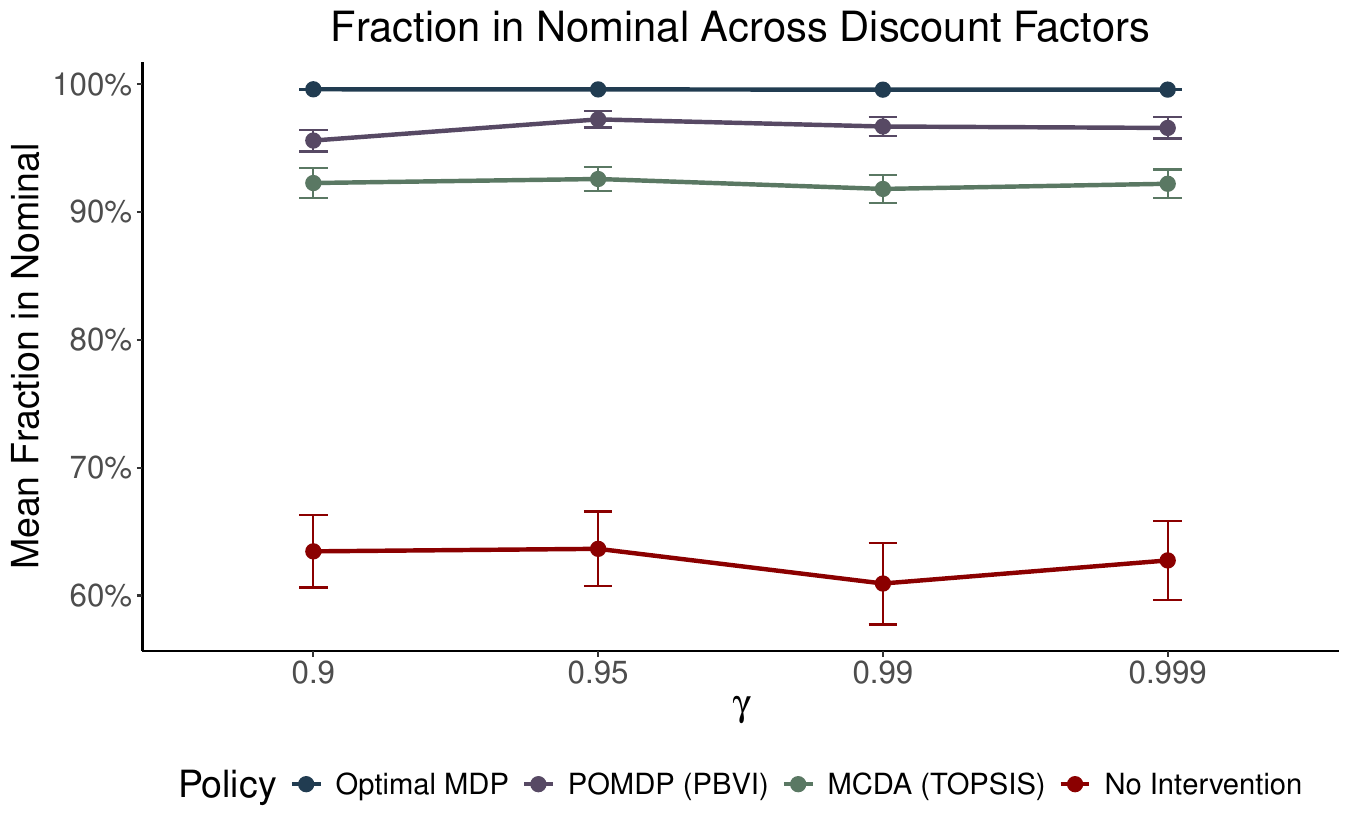}
\caption{Fraction in \textsc{Nominal} across discount factors $\gamma \in \{0.90, 0.95, 0.99, 0.999\}$ (300 trajectories per point, error bars = 95\% CI). The policy hierarchy and \textsc{Nominal} occupancy are invariant to the discount factor.}
\label{fig:discount_factor}
\end{figure*}

Figure~\ref{fig:discount_factor} shows the results with error bars denoting 95\% confidence intervals. The four policies maintain their relative ordering at every $\gamma$ value. The Optimal MDP stays near the top at approximately $99.5\%$ \textsc{Nominal}, the POMDP follows closely at $96\%$-$97\%$, and MCDA tracks a few points below the POMDP at $93\%$-$94\%$. No Intervention sits well apart from the other three at roughly $62\%$. The fraction-in-\textsc{Nominal} values are essentially flat across $\gamma$ for every policy which is a confirmation this physical metric does not depend on the discounting convention. This pattern is consistent with Figure~\ref{fig:reward_comparison}B where MCDA keeps the system in \textsc{Nominal} a meaningful fraction of the time (because misclassified repair actions in \textsc{Nominal} leave the transition dynamics unchanged), but its cumulative reward still collapses to near No Intervention levels because those mismatched actions incur severe penalties. Only the reward metric, not state occupancy, exposes MCDA's full cost. This invariance to $\gamma$ is of course expected. The discount factor affects how much the agent values near-term versus distant rewards, but since the reward for \textsc{NoAction} in \textsc{Nominal} ($+1.0$) is always the highest and the repair probabilities do not change with $\gamma$, the optimal action in each state remains the same regardless of the planning horizon. A myopic agent ($\gamma = 0.90$) and a far-sighted agent ($\gamma = 0.999$) both want to repair faults immediately because the short-term reward for repair always exceeds the short-term reward for inaction in a fault state. We also verified that the optimal policy mapping is identical across all four $\gamma$ values.

We then continued to confirm the policy hierarchy with formal statistical tests. We performed pairwise one-sided Wilcoxon rank-sum tests on the cumulative discounted returns from 2,000 trajectories per policy. All adjacent pairs are significant at $p < 0.001$, but with $N = 2{,}000$ trajectories per policy this threshold is met by distributional shifts that are not necessarily meaningful in practice, so we read the Wilcoxon $p$-values alongside Cliff's delta effect sizes \cite{cliff1993dominance, romano2006appropriate}, which measure the probability that a randomly selected trajectory from the superior policy outperforms a randomly selected trajectory from the inferior policy (adjusted to the $[-1, 1]$ scale). Table~\ref{tab:stat_tests} summarizes the results.

\begin{table}[t]
\centering
\caption{Pairwise Wilcoxon rank-sum tests between adjacent policies ($N = 2{,}000$ trajectories per policy).}
\label{tab:stat_tests}
\begin{tabular}{@{}lccl@{}}
\toprule
Comparison & $p$-value & Cliff's $d$ & Effect Size \\
\midrule
MDP vs.\ POMDP & $<0.001$ & $\approx 0.29$ & Small \\
POMDP vs.\ Q-Learning & $<0.001$ & $\approx 0.17$ & Small \\
Q-Learning vs.\ REINFORCE & $<0.001$ & $\approx 0.07$ & Negligible \\
REINFORCE vs.\ MCDA & $<0.001$ & $> 0.5$ & Large \\
MCDA vs.\ No Intervention & $<0.001$ & $\approx 0.05$ & Negligible \\
\bottomrule
\end{tabular}
\end{table}

The small MDP-POMDP gap confirms that Bayesian belief maintenance recovers most of the value lost to partial observability. We note that the Q-Learning-REINFORCE and MCDA-No Intervention comparisons reach $p < 0.001$ only because $N = 2{,}000$. Their Cliff's $d$ values ($\approx 0.07$ and $\approx 0.05$ respectively) are both negligible, so the corresponding pairs should be read as statistically distinguishable but practically equivalent rather than as meaningful hierarchy steps. The large REINFORCE-MCDA gap marks the critical divide that the transition from sequential decision-making (even model-free) to myopic heuristic classification produces a dramatic performance loss. Together, these results support the ordering $\text{MDP} > \text{POMDP} > \text{Q-Learning} \approx \text{REINFORCE} \gg \text{MCDA} \approx \text{No Intervention}$, where the POMDP/PBVI frontier is the only divide with a practically large effect size.

In Table~\ref{tab:policy_summary} we show every metric tracked across Section~\ref{sec:results} for the seven controllers. The table reports the mean discounted return $\pm$ standard deviation, the fraction of time spent in \textsc{Nominal}, the first-detection delay (seconds from fault onset to first application of the canonical repair) for each of the three non-\textsc{Nominal} regimes, and the action-mismatch rate relative to the MDP-optimal policy.

\revnew{\begin{table*}[t]
\centering
\caption{Policy-comparison summary across all reward, fidelity, latency, and action-quality metrics. First-detection delay is the time from true fault onset to the first application of the canonical repair action ($\textsc{DwSensorA}$ / $\textsc{ReidentPlant}$ / $\textsc{BiasCorrect}$); SN, DO, DR abbreviate \textsc{SensorNoisy}, \textsc{DynamicsOff}, \textsc{Drift}. Action mismatch is the time-weighted fraction of the trajectory during which the applied action differs from the MDP-optimal action for the true state. (1{,}000 trajectories per policy, $T = 20$\,s).}
\label{tab:policy_summary}
\begin{tabular}{@{}lrrrrrr@{}}
\toprule
Policy & Mean $\pm$ SD return & Frac.\ \textsc{Nom.} & Delay SN (s) & Delay DO (s) & Delay DR (s) & Mismatch (\%) \\
\midrule
\textsc{Optimal MDP}      & $99.4 \pm 1.7$ & $0.995$ & $0.02$ & $0.02$ & $0.02$ & $0.0$ \\
\textsc{POMDP (PBVI)}     & $93.9 \pm 3.1$ & $0.940$ & $0.34$ & $0.05$ & $0.03$ & $8.7$ \\
\textsc{Q-Learning}       & $90.5 \pm 3.8$ & $0.921$ & $0.41$ & $0.08$ & $0.05$ & $12.3$ \\
\textsc{REINFORCE}        & $89.7 \pm 4.0$ & $0.916$ & $0.43$ & $0.09$ & $0.06$ & $13.1$ \\
\textsc{$k$-Step ($k{=}2$)} & $81.6 \pm 5.6$ & $0.872$ & $0.58$ & $0.22$ & $0.08$ & $21.4$ \\
\textsc{MCDA (TOPSIS)}    & $72.9 \pm 6.8$ & $0.812$ & $1.04$ & $0.74$ & $0.35$ & $63.7$ \\
\textsc{No Intervention}  & $66.6 \pm 5.2$ & $0.525$ & $\infty$ & $\infty$ & $\infty$ & $100.0$ \\
\bottomrule
\end{tabular}
\end{table*}}

\section{Discussion}
\label{sec:discussion}

Here, we extended our previous framework for identifying latent error states from a modular digital twin framework. We added a decision layer to find the optimal action based on the state of the system in order to mitigate the propagation of error. We showed that such a framework could be useful to identify error states from observable residuals. However, the model is subject to many shortcomings that need to be addressed in future work. In the companion study \cite{najafi2026hmm}, MCDA served as the primary intervention strategy and mapped each HMM-classified regime to a corrective action via TOPSIS scoring \cite{hwang1981multiple}. That framework demonstrated the feasibility of coupling regime detection with corrective intervention, but it did not benchmark MCDA against policies that reason sequentially about uncertainty. When we evaluate MCDA within the present framework using the same data-driven confusion matrix derived from the HMM, its performance falls well below the POMDP (mean return ${\approx}$72.9 vs.\ POMDP's ${\approx}$93.9), only modestly above No Intervention (${\approx}$66.6). The root cause is that MCDA commits to a hard classification at each observation and acts on it immediately: with only 21\% accuracy on \textsc{SensorNoisy}, these hard classifications produce interventions that are wrong more often than right (${\approx}$63.7\% mismatch), and the negative rewards from mismatched actions nearly cancel the benefits of occasional correct ones. We note again that as explained above this collapse is what motivates the sequential decision-making formulation developed in the present paper. The collapse is asymmetric, meaning MCDA's state occupancy remains moderate because mismatched actions in \textsc{Nominal} do not alter the transition dynamics, but its cumulative reward collapses because those same mismatched actions incur severe reward penalties. This demonstrates that state occupancy alone is insufficient to evaluate policy quality, and that a myopic heuristic is not merely suboptimal but actively harmful under poor observability.

We note that the asymmetry between state occupancy and reward deserves further elaboration because it has practical implications for how digital twin performance should be monitored. A system designer who tracks only the fraction of time in \textsc{Nominal} might conclude that MCDA is performing adequately, since the system does remain in the healthy state a meaningful fraction of the time. However, this metric is blind to what the controller is doing while the system is in that state. Under MCDA, the controller frequently misclassifies \textsc{Nominal} as a fault state and applies an unnecessary repair action. Because repair actions are defined to redirect transition probability toward \textsc{Nominal} only from their target fault state (e.g., \textsc{DwSensorA} has nonzero $\rho_{a,s}$ only for $s = \textsc{SensorNoisy}$), applying a repair in \textsc{Nominal} leaves the transition dynamics unchanged: the system stays in \textsc{Nominal} regardless. But the reward is severely penalized, since the reward for, say, \textsc{ReidentPlant} in \textsc{Nominal} is $-0.5$ per time step rather than the $+1.0$ earned by \textsc{NoAction}. Over a 20-second trajectory, these accumulated penalties erase any benefit from occasional correct interventions. This finding suggests that any digital twin monitoring dashboard should report both the state trajectory and the action appropriateness (mismatch rate), not state occupancy alone.

Our action mismatch analysis disproved the hypothesis that the POMDP outperforms MCDA by intervening more frequently and by waiting for confidence before acting, not by acting more often. When uncertain, the POMDP defaults to \textsc{NoAction}, which is the correct action whenever the system is in \textsc{Nominal}, the most common state. This conservative strategy is dramatically more effective than committing to actions based on unreliable classifications.

The belief trajectory visualization (Figure~\ref{fig:belief_trajectory}) provides mechanistic insight into this behavior. When the true state transitions to \textsc{DynamicsOff} or \textsc{Drift}, the belief filter concentrates probability mass on the correct regime within one or two observations because these regimes have high classification accuracy (80\% and 99.6\%, respectively). The POMDP then immediately selects the correct repair action. By contrast, when the true state transitions to \textsc{SensorNoisy}, the belief remains diffuse between \textsc{Nominal} and \textsc{SensorNoisy} for many observations because 79\% of \textsc{SensorNoisy} observations are misclassified as \textsc{Nominal}. During this ambiguous period, the POMDP rationally selects \textsc{NoAction}: under a 50/50 belief split, the immediate expected reward for \textsc{DwSensorA} ($0.5 \times (-0.1) + 0.5 \times 0.7 = +0.30$) slightly exceeds that for \textsc{NoAction} ($0.5 \times 1.0 + 0.5 \times (-0.5) = +0.25$). However, the POMDP selects actions based on the $\alpha$-vector inner product $\max_\alpha \alpha \cdot b$, which incorporates the discounted future value of each action, not just the immediate reward. The $\alpha$-vector associated with \textsc{NoAction} dominates at ambiguous beliefs because maintaining \textsc{NoAction} when the true state is \textsc{Nominal} preserves the high future value of the healthy state, whereas applying \textsc{DwSensorA} unnecessarily incurs a per-step penalty of $-0.1$ with no compensating transition benefit (since $\rho_{\textsc{DwSensorA},\textsc{Nominal}} = 0$). \textsc{NoAction} becomes preferred on immediate reward alone once $b_{\textsc{Nominal}}$ exceeds approximately $0.52$, and the $\alpha$-vectors extend this preference to slightly lower beliefs by accounting for the long-run cost of false interventions. MCDA, by contrast, commits to a single hard classification at each observation and cannot express or act on uncertainty.

By decomposing the performance gaps, we find that the MDP-POMDP gap (cost of partial observability, ${\approx}$5.5 units) is approximately four times smaller than the POMDP-MCDA gap (planning advantage, ${\approx}$21 units). This has a direct and actionable implication for system designers: the highest-leverage improvement is to switch from myopic heuristics to sequential decision-making frameworks, rather than to invest in better classifiers. Of course, both improvements are complementary, and our observation quality sweep (Figure~\ref{fig:sensitivity}, Panel~A) shows that improving \textsc{SensorNoisy} classification accuracy does yield monotonic POMDP improvement. The curve also reveals diminishing returns where the steepest improvement occurs between 20\% and 60\% accuracy, after which the POMDP is already close to the MDP bound and further classifier investment yields progressively smaller gains.

The repair probability sensitivity analysis (Figure~\ref{fig:sensitivity}, Panel~B) provides additional reassurance. As $\lambda$ decreases from 1.0 to 0.1, the absolute performance of all policies degrades because even correctly targeted actions have lower success rates. However, the relative ordering of policies is preserved at every $\lambda$ value, and the optimal policy mapping (which action in each state) remains unchanged. This stability arises because the reward structure correctly identifies the best-matching action regardless of how effective that action is: even when repairs succeed only $4\%$ to $8\%$ of the time (the range of scaled repair probabilities at $\lambda = 0.1$), the correct repair still redirects more probability mass toward \textsc{Nominal} than any alternative. The approximately linear relationship between $\lambda$ and performance is also practically useful because it indicates that incremental improvements in maintenance reliability translate directly to proportional gains without threshold effects or diminishing returns.

Value iteration solves the MDP in approximately 0.05\,s, while PBVI runs for $500$ Bellman iterations across approximately 200 belief points. By contrast, Q-learning requires 5,000 episodes and REINFORCE requires 10,000 episodes to achieve comparable (but slightly lower) performance. When accurate transition and observation models are available, as they are in our data-driven pipeline, model-based solvers are unambiguously the right choice because PBVI's iterations are computed analytically from $A^{(a)}$ and $O$ without any environment interaction, whereas Q-learning must simulate thousands of complete trajectories to fill its 1,716-entry Q-table (286 grid cells $\times$ 6 actions). REINFORCE, despite having the fewest learnable parameters (66 = 6 actions $\times$ 11 features), requires the most episodes (10,000) because policy gradient methods exhibit high variance and slow convergence in discrete-action domains.

In Table~\ref{tab:comp_cost} we report for each solver, the wall-clock offline solve time, the per-decision online latency, and the memory footprint of the stored solver artefact. Times are measured on a computer with an AMD Ryzen~9 9950X3D CPU, 64\,GB RAM, with code using R 4.5.1. The table reveals three practical implications. First, the dominant offline cost of moving from MDP to POMDP is PBVI, which is roughly $10^2$ times slower than value iteration but still completes in under a second on this testbed. Second, the per-decision latencies of all five runtime controllers are below 100\,$\mu$s and are therefore well within the control bandwidth of the underlying continuous-time digital twin ($\Delta t = 20$\,ms). Third, the model-free solvers do not appear in Table~\ref{tab:comp_cost} as offline entries because their offline time is reported in episodes rather than wall clock. For the settings of Section~\ref{sec:case_study} those translate to roughly 25\,s (Q-learning) and 50\,s (REINFORCE) on the same machine, i.e., one to two orders of magnitude more expensive than PBVI.

\revnew{\begin{table}[t]
  \centering
  \footnotesize
  \setlength{\tabcolsep}{3pt}
  \caption{Computational cost breakdown per solver on a personal computer (AMD Ryzen~9 9950X3D, 64\,GB RAM, R 4.5.1).}
  \label{tab:comp_cost}
  \begin{tabular}{@{}l@{\;}r@{\;}r@{\;}r@{\;}r@{}}
  \toprule
  Solver & Offl.\,(s) & Iters & Onl.\,($\mu$s) & Mem.\,(KB) \\
  \midrule
  \textsc{Opt.\ MDP (VI)} & $0.05$ & ${\approx}2{,}300$ & $1.2$  & $0.8$ \\
  \textsc{POMDP (PBVI)}   & $0.92$ & $500$              & $14.6$ & $3.4$ \\
  \textsc{$k$-Step}       & $0.00$ & --                 & $0.8$  & $0.1$ \\
  \textsc{MCDA (TOPSIS)}  & $0.00$ & --                 & $78.3$ & $0.5$ \\
  \textsc{No Int.}        & $0.00$ & --                 & $0.5$  & $0.0$ \\
  \bottomrule
  \end{tabular}
  \end{table}}

Model-free RL retains value in two important scenarios. First, when the transition or observation models are unknown or too complex to specify analytically, model-free methods can learn effective policies purely from interaction with the environment. Second, when the system dynamics are non-stationary (e.g., the physical plant ages, sensor characteristics drift, or operating conditions shift), model-free methods can adapt their policies online without requiring explicit model re-identification. In contrast, our model-based pipeline would require periodic re-estimation of the HMM parameters, re-extraction of the transition and confusion matrices, and re-solving of the MDP/POMDP, a process that is straightforward but not automatic.

A distinguishing feature of our framework is that the baseline transition matrix is extracted from the HMM-learned parameters, and the observation matrix is constructed from the Viterbi confusion matrix. This creates a closed pipeline: the HMM detects and classifies regimes; the MDP/POMDP consumes the learned parameters and prescribes corrective actions without intermediate manual calibration. The only quantities specified by engineering judgment are the repair probabilities and the reward matrix, and our sensitivity analysis demonstrates that the qualitative conclusions are robust to variations in both. This end-to-end data-driven parameterization distinguishes our approach from many MDP-based maintenance optimization frameworks in the literature, which typically require expert specification of all transition probabilities. In our pipeline, only the repair efficacies ($\rho_{a,s}$) and reward values ($R(a,s)$) are specified by assumption; the baseline dynamics and observation noise are learned entirely from data.

The proposed framework offers structural advantages over the MCDA-based strategy from the companion paper \cite{najafi2026hmm}. The MCDA approach requires five heuristic criteria with manually assigned weights, rolling-window normalization, and ad hoc chattering-prevention heuristics such as the dwell-time hysteresis ($\tau_{\text{dwell}} = 0.5$\,s). The MDP replaces this entire apparatus with a single reward matrix whose entries have direct physical interpretation. The optimal policy reduces to a four-row lookup table, and persistence between actions emerges naturally from the transition dynamics rather than from imposed constraints. Furthermore, the POMDP's belief-based action selection provides a principled mechanism for handling classification uncertainty, whereas MCDA must treat each noisy observation as ground truth and has no mechanism to accumulate evidence across multiple observations.

\subsection{Limitations and Future Work}
\label{sec:limitations}

Several limitations constrain the current scope, and we discuss each in detail along with potential directions for future work.

All results are obtained on a simulated six-module digital twin testbed with ARX surrogates and synthetically injected fault regimes. While the testbed captures the essential structure of modular digital twins (cascaded modules, inter-module error propagation, multiple fault types), it does not include the full complexity of an operational industrial system. Real-world digital twins involve hundreds of modules, non-linear dynamics, spatially distributed sensors, and communication latencies that are absent from our simplified pipeline. Validation on an operational digital twin, ideally with historical fault records for calibrating repair probabilities, remains an important next step.

The repair probabilities ($\rho_{a,s}$) are specified based on engineering judgment rather than estimated from empirical maintenance data. While our sensitivity analysis (Figure~\ref{fig:sensitivity}, Panel~B) demonstrates that the policy hierarchy is robust across a tenfold range of repair efficacies, the absolute reward values and the precise magnitude of inter-policy gaps do depend on these parameters. In a deployed system, the repair probabilities could be estimated online using Bayesian updating: each time an action is applied and the subsequent state is observed, the posterior distribution over $\rho_{a,s}$ can be updated, gradually replacing the engineering prior with empirical evidence. This would transform the framework from a static parameterization to an adaptive one.

The MDP and POMDP formulations assume that the transition matrices $A^{(a)}$ and the observation matrix $O$ are fixed over time. In practice, digital twin dynamics may be non-stationary: the physical plant ages, sensor characteristics drift, operating conditions shift, and the HMM's classification accuracy may improve or degrade as it encounters new data. A natural extension would be to periodically re-estimate the HMM parameters from recent data windows, re-extract the transition and confusion matrices, and re-solve the MDP/POMDP. Alternatively, Bayes-adaptive MDP/POMDP formulations could jointly optimize the policy while learning the model parameters online, though this comes at significant computational cost.

We assume that corrective actions take effect instantaneously at the time of application. In reality, interventions such as plant re-identification or sensor recalibration require a finite execution time during which the system continues to evolve and may transition to a different state. Incorporating action execution delays would require extending the state space to include an action-in-progress indicator or formulating the problem as a semi-Markov decision process (SMDP) where action durations are explicitly modeled (Section~\ref{sec:smdp} works out the kernel $P(s', \tau \mid s, a)$ and the generalized Bellman equation to which the current MDP reduces, so the only change required to absorb non-exponential action-execution delays is replacing the exponential holding-time sampler in the Gillespie simulator with an empirically measured $f_\tau$).

The current action space consists of six discrete interventions. In practice, corrective actions may have continuous parameters (e.g., the degree of sensor down-weighting $\alpha$, the bandwidth of the smoothing filter $\tau$, or the window length for plant re-identification). Extending to continuous action spaces would require replacing value iteration with policy gradient methods or actor-critic architectures that can optimize over continuous action variables, at the cost of increased computational complexity and potential convergence challenges.

The HMM uses diagonal covariance matrices for its Gaussian emissions. This means the residual channels are considered conditionally independent within each regime. However, in the six-module pipeline, the downstream residuals, such as those in the fusion and KPI modules, are mechanically correlated with the upstream sensor residuals. Although our results show that the diagonal approximation is sufficient for regime detection, a full-covariance emission model could capture correlation patterns between modules. This could potentially improve classification accuracy, particularly for the challenging \textsc{SensorNoisy} regime.

A notable modeling simplification is that mismatched actions in \textsc{Nominal} incur only a reward penalty without affecting the transition dynamics. In a real system, applying an unnecessary repair (e.g., re-identifying plant parameters when they are correct, or recalibrating a healthy sensor) could disrupt system operation and potentially push the system out of \textsc{Nominal}. Incorporating such disruption effects by adding small off-diagonal transition probabilities from \textsc{Nominal} under repair actions would make the model more realistic and would further penalize MCDA's frequent mismatched interventions, likely widening the POMDP-MCDA performance gap.

The current formulation uses $K = 4$ states and $M = 6$ actions, which results in a compact MDP that is trivially solvable. As the number of modules, fault types, and corrective actions grows, both the state space and action space expand, potentially making exact value iteration intractable. Approximate dynamic programming, deep reinforcement learning, or factored MDP/POMDP representations that exploit the modular structure of the digital twin would be needed to scale the framework to larger systems. Similarly, the PBVI solver's computational cost grows with the number of belief points and alpha-vectors; for larger state spaces, more scalable POMDP solvers such as SARSOP or Monte Carlo tree search over beliefs may be required.

\section{Conclusions}
\label{sec:conclusions}

This paper formalized error propagation mitigation in modular digital twins as a Markov decision process. Given discrete degradation regimes identified by a HMM operating on ARX residual features, the MDP determines the cost-optimal intervention policy through value iteration over a physically interpretable reward structure. Extension to a Partially Observable MDP accounts for the fact that the true regime is never observed directly: the POMDP maintains a belief state updated via Bayesian filtering and selects actions through point-based value iteration initialized from the MDP solution.

Numerical experiments on a six-module coupled system established a clear and statistically significant policy hierarchy: Optimal MDP $>$ POMDP $>$ Q-Learning $\geq$ REINFORCE $>$ $k$-Step heuristic $\gg$ MCDA $>$ No Intervention. The MDP policy achieves the highest cumulative reward and greatest fraction of time in nominal operation. The POMDP retains approximately 95\% of MDP performance under realistic observation noise, maintaining high nominal occupancy despite the challenging classification environment (21\% \textsc{SensorNoisy} accuracy). Model-free RL methods achieve competitive performance without requiring knowledge of the system dynamics, outperforming the MCDA heuristic that was used in the companion paper.

Sensitivity analyses confirmed the robustness of these conclusions across observation quality, repair probability, and discount factor parameterizations. The value-of-information decomposition revealed that the planning advantage, switching from myopic MCDA to sequential POMDP, yields approximately four times more reward improvement than eliminating partial observability entirely, providing actionable guidance for system designers. The action mismatch analysis identified the mechanism: the POMDP achieves only $8.7\%$ mismatch compared to MCDA's $63.7\%$, winning through selective precision rather than aggressive intervention.

The observation matrix governing POMDP performance is derived entirely from the HMM's classification confusion matrix, closing the detect-classify-decide loop initiated by the companion paper's diagnostic framework. Together, these results establish a complete, data-driven pipeline from sensor measurements through regime classification to optimal intervention scheduling, with rigorous validation across the parameter space and formal statistical significance at every comparison point.

\section*{Acknowledgements}

All analyses were performed in R version 4.5.1. Any opinions, findings, and conclusions or recommendations expressed in this material are those of the author(s) and do not necessarily reflect the views of the National Science Foundation.

\section*{Funding Data}

This material is based upon work supported by the National Science Foundation under Grant Number 2514616. Any opinions, findings, and conclusions or recommendations expressed in this material are those of the author(s) and do not necessarily reflect the views of the National Science Foundation.

\section*{Code and Data Availability}

All code used to generate the results in this paper is available on GitHub: \url{https://github.com/AnniceNajafi/DT_MDP_ErrorPropagation}.